%% file: main.tex
\input{settings}

\title{Retrieval of Surface Solar Radiation through Implicit Albedo Recovery from Temporal Context}

\author{Yael Frischholz$^1$, Devis Tuia$^{2}$, Michael Lehning$^{1}$}
\date{
    $^1$Laboratory of Cryospheric Sciences, Ecole Polytechnique Fédérale de Lausanne, Sion, Switzerland.\\
    $^2$Environmental Computational Science and Earth Observation Laboratory, Ecole Polytechnique Fédérale de Lausanne, Sion, Switzerland.
}

\begin{document}
\maketitle
\input{0_abstract}
\newpage
\input{1_intro}
\input{2_SSR}
\input{3_methods}
\input{4_results}

\input{5_discussion}

\input{6_conclusion}
\input{ack_data_cred}

\clearpage
\bibliography{references_arxiv}

\thispagestyle{empty}
\appendix
\input{appendix}
\end{document}

%% file: settings.tex
\documentclass{article}
\usepackage[T1]{fontenc}
\usepackage[utf8]{inputenc}

\usepackage{fourier}
\usepackage[cmex10]{amsmath} 

\usepackage{titlesec, blindtext, color}
\usepackage{pifont}
\newcommand{\cmark}{\ding{51}}
\newcommand{\xmark}{\ding{55}}
\definecolor{gray75}{gray}{0.75}
\newcommand{\hsp}{\hspace{20pt}}
\titleformat{\chapter}[hang]{\Huge\bfseries}{\thechapter\hsp\textcolor{gray75}{|}\hsp}{0pt}{\Huge\bfseries}
\usepackage{booktabs}

\usepackage{pdfpages}
\usepackage{graphicx}
\graphicspath{ {figures/} }
\usepackage{wrapfig}
\usepackage{lscape}
\usepackage[counterclockwise, figuresleft]{rotating}
\usepackage[labelfont=it, textfont=it]{caption}
\usepackage{svg}
\usepackage{subcaption}
\usepackage{pgfplots}
\usepackage{float}
\pgfplotsset{width=10cm,compat=1.9}
\usepgfplotslibrary{external}
\usepackage{booktabs}
\usepackage{microtype}
\usepackage{url}


\usepackage{makecell}
\usepackage{multirow}
\usepackage{textcomp}
\usepackage{hyperref}
\hypersetup{pdfborder={0 0 0},
	colorlinks=true,
	linkcolor=blue,
	citecolor=blue,
	urlcolor=blue}

\numberwithin{table}{section}
\numberwithin{equation}{section}
\numberwithin{figure}{section}
\usepackage[bottom]{footmisc}

\usepackage{epstopdf}
\usepackage{geometry}
\usepackage{setspace} 
\usepackage{titlesec}

\usepackage{siunitx}
\usepackage{blindtext} 

\usepackage{amssymb}
\usepackage{empheq}
\usepackage{mathtools}

\usepackage{xcolor}

\usepackage{fancyhdr}
\usepackage{sidecap}

\usepackage{natbib}


%% file: 0_abstract.tex
\begin{abstract}
Accurate retrieval of surface solar radiation (SSR) from satellite imagery critically depends on estimating the background reflectance that a spaceborne sensor would observe under clear-sky conditions. Deviations from this baseline can then be used to detect cloud presence and guide radiative transfer models in inferring atmospheric attenuation. Operational retrieval algorithms typically approximate background reflectance using monthly statistics, assuming surface properties vary slowly relative to atmospheric conditions. However, this approach fails in mountainous regions where intermittent snow cover and changing snow surfaces are frequent.
We propose an attention-based emulator for SSR retrieval that implicitly learns to infer clear-sky surface reflectance from raw satellite image sequences. Built on the Temporo-Spatial Vision Transformer, our approach eliminates the need for hand-crafted features such as explicit albedo maps or cloud masks. The emulator is trained on instantaneous SSR estimates from the HelioMont algorithm over Switzerland, a region characterized by complex terrain and dynamic snow cover. Inputs include multi-spectral SEVIRI imagery from the Meteosat Second Generation platform, augmented with static topographic features and solar geometry. The target variable is the HelioMont's SSR, computed as the sum of its direct and diffuse horizontal irradiance components, given at a spatial resolution of 1.7km. We show that, when provided a sufficiently long temporal context, the model matches the performances of albedo-informed models, highlighting the model's ability to internally learn and exploit latent surface reflectance dynamics. 
Our model, namely HeMu, achieves root-mean-square deviations from ground truth of 134.0 ~W/m\textsuperscript{2} for instantaneous measurements, 52.9 ~W/m\textsuperscript{2} and 22.5 ~W/m\textsuperscript{2} for daily and monthly aggregates respectively, matching the accuracy of HelioMont's products, with substantial computational speedup over the original CPU-based pipeline.
Our geospatial analysis shows this effect is most powerful in mountainous regions and improves generalization in both simple and complex topographic settings.
Code and data sets are publicly available: \texttt{\url{https://github.com/frischwood/HeMu-dev.git}}
\end{abstract}

%% file: 1_intro.tex
\section{Introduction}

Accurate estimates of surface solar radiation (SSR) are essential for a wide range of climate, energy, and environmental applications.
Notably, as electricity generation systems across the world transition to more renewable sources \citep{iea_renewables_2024}, high resolution maps of solar climatology at regional to global scales are key in the site selection process of solar photovoltaic power plants. 
Moreover, historical gridded solar resource data, serving as a primary input for many recent data-driven SSR forecast models (e.g. \citep{paletta_advances_2023,sehrawat_solar_2023,al-dahidi_enhancing_2024,carpentieri_extending_2025, carpentieri_data-driven_2024}), has become a valuable asset. 

Geostationary satellite-based imagery allows to retrieve SSR estimates that have outperformed ground-station interpolations from sparse  measurements for more than two decades already \citep{perez_comparing_1997, ineichen_derivation_1999}.
Space-borne multi-spectral images enable SSR retrieval by capturing top-of-atmosphere reflectance in various regions of the spectrum, from which atmospheric attenuation is inferred to estimate the irradiance fraction that reaches the surface. This inversion problem requires disentangling atmospheric effects from surface reflectance \citep{huang_estimating_2019}. 
State-of-the-art SSR retrieval algorithms (e.g. CAMS \citep{schroedter-homscheidt_surface_2022}, SARAH-3 \citep{pfeifroth_sarah-3_2024}, HelioMont \citep{castelli_heliomont_2014} etc.) address this by estimating the reflectance that would be observed under clear-sky conditions, referred to as background or clear-sky reflectance. For a geostationary platform, background reflectance is primarily a function of the solar incident angle and the surface albedo \citep{cano_method_1986, huang_estimating_2019}.
To estimate this value, methods typically maintain reflectance statistics based on a rolling buffer of recent measurements, under the assumption that surface properties evolve more slowly than atmospheric conditions. Deviations from this baseline are attributed to cloudiness \citep{cano_method_1986}. 
Radiative transfer models (RTMs), using ancillary data on atmospheric states (e.g., aerosols, humidity), are then employed to correct for atmospheric absorption and scattering \citep{emde_libradtran_2016, lefevre_mcclear_2013}.

This approach however faces challenges in regions with complex topography that experience periods of intermittent snow cover. Rapid changes in snow surface properties are also problematic. For affected pixels, commonly used monthly background reflectance statistics become obsolete due to daily to hourly changes. Additionally, in the same regions, the spectral similarity of snow and clouds complicates cloud masking routines and, depending on the spatial resolution of the images, an ortho-rectification may be required to correct optical deformations. \citep{carpentieri_satellite-derived_2023} demonstrated that these limitations were especially critical in Alpine regions above 1000 m asl. during winter, confirming the significant role of snow albedo in the generalization of SSR retrievals across regions with varying topographic complexity.

Recently, several machine learning (ML)-based models for SSR retrieval from satellite imagery were proposed (e.g. \citep{jiang_deep_2019,lu_predicting_2023, gurel_state_2023, schuurman_surface_2024}). They are either trained to emulate physics-based RTMs or directly use ground station measurements as target. The advantage of the former lies in the availability and consistency of the usually gridded, target data. The latter offers more accurate, but sparse, point target data.  

Most relevant to the present work is the emulator from \citep{schuurman_surface_2024}, which adopted a convolution-based architecture to emulate instantaneous SSR estimates of SARAH-3 \citep{pfeifroth_sarah-3_2024} and fine-tuned the estimates on various ground station measurements. This approach outperformed the base model and improved out-of-domain generalization. 

In this context, \citep{verbois_retrieval_2023} pointed out that ground albedo was a key modulator of the estimate accuracy and generalization.  

While the previous ML-based methods focused on mapping concurrent observations to SSR estimates, \citep{carpentieri_data-driven_2024} first introduced temporal context to estimate SSR. Although not using satellite images as primary input, the authors showed that context-aware models (6 time steps) yielded better SSR estimates when provided with a sufficiently large spatial receptive field, and suggested that this context implicitly allowed the model to learn the advective dynamics of clouds. 

In this article, we show that providing a temporal context, a window of past satellite observations, significantly improves ML-based SSR retrievals by enabling the model to implicitly reconstruct background reflectance. Specifically, we show that a context-aware attention-based model can match the performance gains of albedo-informed models, which largely improve the generalization of SSR estimates to mountainous regions, without requiring surface albedo as an input. Our findings suggest that context acts as a soft memory for surface conditions, echoing the compositing logic used in physical algorithms, but implemented entirely via learned attention. 

According to these findings, we present HeMu, a deep learning emulator for SSR retrieval based on the Temporo-Spatial Vision Transformer (TSViT) \citep{tarasiou_vits_2023}, a relatively light-weight architecture. Trained on high-resolution SSR estimates from the physics-based HelioMont method \citep{castelli_heliomont_2014}, HeMu achieves state-of-the-art accuracy over Switzerland, particularly in snow-affected and mountainous regions. It further provides substantial computational speedup compared to the RTM-based method it emulates.

The rest of the article is structured as follows: Section \ref{sec:ssr} provides some background on SSR retrieval from satellite imagery. Section \ref{sec:methodology} presents the methods, including model architecture, data and experiments. Section \ref{sec:results} presents the results discussed in Section \ref{sec:discussion}. 

%% file: 2_SSR.tex
\section{Background on SSR Retrieval from Satellite Imagery}
\label{sec:ssr}

Top-of-atmosphere (TOA) reflectance, as measured by satellite-borne multispectral sensors, integrates the effects of atmospheric scattering and surface reflection. Accurate retrieval of SSR involves disentangling these components to estimate the downwelling shortwave radiation at the surface \citep{huang_estimating_2019}.

Physics-based approaches model the atmospheric contribution using RTMs. An example is the libRadTran \citep{emde_libradtran_2016} model, which simulates radiance given atmospheric composition (e.g., ozone, water vapor, aerosols). These models often operate via look-up tables (LuTs) that map TOA reflectance and ancillary atmospheric data to SSR: $\mathrm{LuT}(\rho_{\mathrm{TOA}}, \text{ atmospheric state}) \rightarrow \mathrm{SSR}$. While conceptually robust, such methods critically depend on the quality and availability of auxiliary data.

In contrast, statistical methods exploit empirical relationships between cloudiness and SSR. The canonical Heliosat method \citep{cano_method_1986} quantifies cloud presence via the \emph{cloud index}:
\[
n = \frac{\rho - \rho_{\mathrm{cs}}}{\rho_{\max} - \rho_{\mathrm{cs}}}
\]
where, $\rho$ is the observed reflectance, $\rho_{\mathrm{cs}}$ is the background (clear-sky) reflectance, and $\rho_{\max}$ is the upper bound reflectance measured over full cloud cover. The cloud index is used to estimate the clear-sky index :
\begin{align*}
k_T^* = \frac{G}{G_{cs}} \approx 1-n
\end{align*} 
where, $G$ is the actual SSR and $G_{\mathrm{cs}}$ is the SSR under clear-sky conditions \citep{hammer_solar_2003}. The latter is typically computed using empirical clear-sky models (e.g.\citep{bird_simple_1986, muneer_solar_1990, hay_calculating_1993}).

Contemporary methods, such as SARAH-3 \citep{pfeifroth_sarah-3_2024} and HelioMont \citep{castelli_heliomont_2014}, adopt hybrid strategies: they retain the statistical Heliosat structure for estimating $k_T^*$ while using RTMs to derive $G_{\mathrm{cs}}$. 

\subsection{Challenges in Snowy and Mountainous Terrain}
Accurate SSR retrieval is particularly challenging in regions with seasonally variable snow cover and complex topography. In such conditions, traditional approaches systematically underestimate SSR during snow-covered periods and overestimate it otherwise \citep{carpentieri_satellite-derived_2023}. This stems from the difficulty of estimating background reflectance ($\rho_{\mathrm{cs}}$) under cloudy conditions over dynamic snow surfaces.

Heliosat-style methods typically estimate $\rho_{\mathrm{cs}}$ and $\rho_{\max}$ from reflectance statistics over a monthly window, using the 5th and 99th percentiles, respectively. In transitional snow periods, e.g. near snow lines or during the winter transition periods, this method can lead to underestimation of $\rho_{\mathrm{cs}}$, resulting in an inflated cloud index and a negatively biased SSR estimate.

Moreover, mountainous regions introduce additional complexity. High-resolution imagery suffers from geometric distortions due to satellite viewing angles, leading to spatially varying pixel footprints. These orographic deformations impact the fidelity of the sensed reflectance. Furthermore, surface reflectance is anisotropic \citep{tan_illumination_2010,von_rutte_how_2021} and depends on illumination geometry (incident angle); thus, identical SSR levels can yield different TOA reflectance depending on terrain orientation.

\subsection{The HelioMont Method}
HelioMont \citep{castelli_heliomont_2014} is a state-of-the-art SSR retrieval method tailored to Alpine regions. It addresses the challenges of snow dynamics and complex topography via three key elements:

\begin{enumerate}
    \item \textbf{Snow-Cloud Discrimination}: HelioMont incorporates the SPARC algorithm \citep{khlopenkov_sparc_2007} to produce a cumulative and continuous score based on multi-spectral features and from which snow/cloud masks are derived.  
    SPARC improves robustness to unavailable data and better treats mixed-phase conditions compared to the more common classification tree-based algorithms.

    \item \textbf{Time-Resolved Background Reflectance}: Rather than relying on monthly reflectance percentiles, HelioMont reconstructs hourly clear-sky reflectance time series. Using a 10-day archive of pixels identified as clear-sky (via SPARC), it composites dynamic $\rho_{\mathrm{cs}}$ fields that better capture diurnal and sub-seasonal variability, especially near the snow line. More recent observations are prioritized by weighting them relatively to their acquisition time.

    \item \textbf{Topographic Correction}: HelioMont orthorectifies reflectance using the 1-arcsecond SRTM DEM \citep{opentopography_shuttle_2013}, compensating for terrain-induced distortions. It also takes into account the local illumination geometry to correct for slope and aspect effects on observed reflectance \citep{tan_illumination_2010}. These steps mitigate both spatial deformation and radiometric bias.
\end{enumerate}

HelioMont provides direct and diffuse SSR components corrected for horizon effects, which are very effective in mountainous terrains \citep{carpentieri_satellite-derived_2023}. However, its applicability outside of Switzerland remains to be demonstrated. 
For a complete description, we refer the reader to the HelioMont technical documentation \citep{stockli_heliomont_2022}.

%% file: 3_methods.tex
\section{Methodology}
\label{sec:methodology}

\subsection{Problem Formulation}
\label{subsec:emulatingHelioMont}
We emulate the HelioMont SSR retrieval algorithm with a supervised deep learning model, which maps a temporal sequence of satellite images and auxiliary data to the SSR field at the final time step:
\[
\mathbf{f}: [\mathbf{x}_{t-T}, \ldots, \mathbf{x}_{t}] \rightarrow \mathbf{y}_t, \quad \mathbf{x}_t \in \mathbb{R}^{H \times W \times C},\quad \mathbf{y}_t \in \mathbb{R}^{H \times W}, 
\]
where $H$ and $W$ are the height and width of the original image, respectively, and $C$ is the number of spectral channels augmented with some auxiliary data.
Here, $\mathbf{y}_t$ is the target SSR, at time $t$, while $\mathbf{x}_{t-T}, \ldots, \mathbf{x}_t$ is a sequence of satellite images and auxiliary features. 
We hypothesize that increasing the temporal context $T$ allows the model to better infer the latent clear-sky background reflectance, similar to the explicit temporal compositing used in HelioMont.

\subsection{Architecture}
\label{subsec:Architecture}
Our model is based on the Temporo-Spatial Vision Transformer (TSViT) architecture for satellite image time series \citep{tarasiou_vits_2023}, a dual-encoder Vision Transformer (ViT) that factorizes temporal and spatial attention to manage the computational complexity of long sequences. TSViT was originally proposed for the segmentation of crop types.

Unlike natural video data, where spatial coherence is dominant, satellite image classification benefits from a \emph{temporo-spatial} approach, in which the evolution of a single pixel's multi-spectral signature carries more information than standard RGB videos \citep{tarasiou_vits_2023}. 

\begin{figure}[H]
    \centering
    \includegraphics[width=0.8\linewidth]{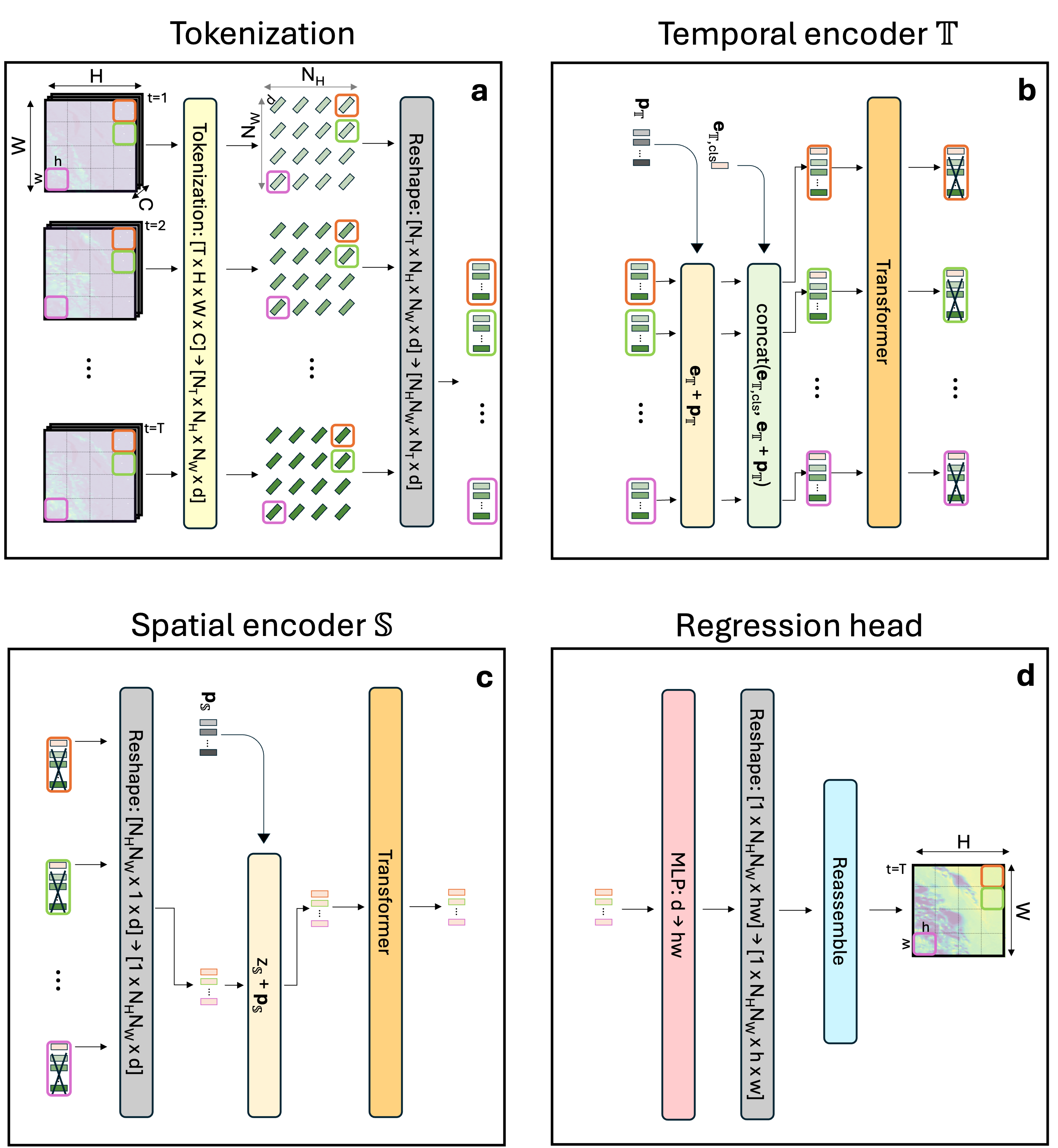}
    \caption{Architecture of the TSViT-r, adapted from \citep{tarasiou_vits_2023}.The input timeseries of length T and C channels, including satellite images, topography-derived features and sun angles, is first tokenized into $N_H \times N_W$ sequences of $N_T$ tokens of dimension d (sub-figure \textbf{a}). Then, the temporal sequences of all patches are fed to the temporal encoder, which is a standard ViT (sub-figure \textbf{b}). The ViT prepends an additional token, $\mathbf{e}_{\textit{T},cls}$, to the representation of every pixel and encodes the sequence through the Transformer. The encoded additional token of each temporal sequence is subsequently passed to the spatial encoder, which is similar to the previous ViT, but does not prepend any additional token (sub-figure \textbf{c}). 
    Finally, the sequence encoded by the spatial Transformer is projected and reshaped back to the original image dimensions (sub-figure \textbf{d}). Further details on the TSViT-r architecture are provided in append \ref{app:archi}.}
    \label{fig:archi_model}
\end{figure}

We therefore encode the spatio-temporal evolution of the background reflectance based on the spectral signature of each pixel. We name our model TSViT-r. As for the original TSViT, we proceed in a sequential manner by factorizing along the temporal and then spatial dimensions with two separate encoders (Fig.~\ref{fig:archi_model}):
\begin{itemize}
\item The original images are devided into $N_H \times N_W$ patches. Each patch is then tokenized along the temporal dimension, so that each embedding $\mathbf{e}_\mathbb{T}$ represents a region of the image across time.
\item In the \textbf{temporal encoder} ($\mathbb{T}$),  a standard positional embedding $\mathbf{p}_\mathbb{T}$ is then added to each token, in order to conserve the temporal ordering among the time-steps. An extra token $\mathbf{e}_{\mathbb{T},\text{cls}}$ is appended to each embedding in the sequence and used as an input to the temporal transformer. After learning, only the extra token is retained and the rest of each sequence is discarded. The retained token should contain all temporal relationships in the temporal domain.
\item In the \textbf{spatial encoder} ($\mathbb{S}$), only the tokens $\mathbf{e}_{\mathbb{T},\text{cls}}$, temporally encoded, are used. Since each token corresponds to a region of the original image (highlighted as red, green and pink outlines), this new transformer only considers spatial relations. As for the temporal encoder, a standard positional embedding $\mathbf{p}_\mathbb{S}$ is added, in order to conserve information about the spatial position of each token in the original image. A standard vision transformer then learns one embedding per patch.
\item All the patch-specific embeddings of dimension $\mathbf{d}$ are used as inputs to a standard MLP in the \textbf{regression head}, which calculates the SSR values for each pixel.
\end{itemize}
The corresponding formalized operations are provided in Appendix ~\ref{app:archi}. Further details on the TSViT backbone are provided in Appendices~\ref{app:vit_background} and \ref{app:TSVIt}.

\subsection{Data}
\label{subsec:data}
The input features consist in satellite images, solar angles and descriptors of the topography. All data used as input features and target is publicly available. A comprehensive list of the input and target features is provided in Table \ref{tab:featuresList}. 

\paragraph{SSR Targets:}
The model is trained to predict instantaneous SSR computed as the sum of HelioMont's direct and diffuse horizontal irradiance products that exclude terrain shadowing and reflections. The dataset spans five years (2015–2020) over Switzerland. The study area bounding box is: [5.75°E-10.75°E] $\times$ [45.75°N-47.88°N] (WGS84). Only the instantaneous on-the-hour timestamps are selected, as opposed to the standard hourly averaged product. 
The spatial grid resolution is 0.05° (WGS84), or $\approx$ 1.7km in the local planimetric reference system (CH1903+). The data was provided by MeteoSwiss. The hourly averaged version is accessible via the MeteoSwiss Open Data platform.

\paragraph{Satellite Inputs:}
Multi-spectral images from the MSG-SEVIRI sensor \citep{schumann_msg_2002} are used. The HRV channel and all infrared bands are selected. One sample per hour, the closest to the round hour, is selected. The image timestamp does not exactly correspond to the scanning time since an entire scan lasts about 12 minutes. As the target uses the same satellite data, there is actually no shift between input and target data. For the evaluation against ground truth measurement however, a shift cannot be avoided. To minimize the shift, we use a 10-minute averaged ground truth (see below). 
The HRV channel's grid resolution corresponds to the target's resolution and justifies the selection of this channel instead of the lower resolution VIS006 and VIS008 visible channels. The data was accessed on the EUMETSAT DataStore via the EUMDAC API. 

\paragraph{Auxiliary Inputs:}
Solar zenith and azimuth angles are computed using \citep{grena_five_2012} for a single year and used for the other years by the hour-of-the-year. Elevation, slope, aspect (as sine and cosine), and a correction factor for illumination ($f_{corr}$) \citep{muller_grid-_2005} are derived from the NASADEM \citep{opentopography_nasadem_2021} using the HORAYZON set of algorithms \citep{steger_horayzon_2022}. HORAYZON provides direct access to the NASADEM via the online NASA Data Pool. The $f_{corr}$ factor includes corrections for the solar incident angle and the shading of the topography. It is defined in further details in Appendix \ref{appendix:fcorr}. $f_{corr}$ is computed as a function of the elevation grid and the solar angles and therefore, as for solar angles, is computed for a single year, assuming negligible inter-annual variations.

\paragraph{Ground Truth and Diagnostics:}
For the quality assessment of the emulator's estimates, measurements from 87 ground-stations of the SwissMetNet \cite{noauthor_automatic_nodate} network are used. The available measurement is the 10-minutes average SSR, directly used for comparison of instantaneous estimates as described above. 

Additionally, HelioMont's internal albedo ($\alpha$) and clear-sky index ($k_T^*$) time-series are used to classify the concurrent pixels' atmospheric and ground reflectance properties. 

\paragraph{Preprocessing:}
All datasets are temporally and spatially aligned to the HelioMont (target) grid. Lower resolution grids are simply up-sampled with a finer grid. Higher resolution grids are down-sampled averaging pixels contained in the target grid cells. Raw SEVIRI data is preprocessed using Satpy \citep{martin_raspaud_pytrollsatpy_2025}. Inputs are stored in NetCDF format. Time steps with high solar zenith angles ($>80^\circ$ anywhere in the study area) are filtered out to avoid large terrain effects. The filtering of solar zenith angles leads to a variation in data points per day over the year (Appendix \ref{appendix:dataset}). Statistics ($\mu$, $\sigma$) of each variable are precomputed and used for normalization. 

\input{table1} 

\subsection{Temporal Context Experiment}
We test the hypothesis that increasing the temporal context $T$ improves SSR estimation by helping the model to implicitly infer background reflectance. Models are trained with context lengths from $T=1$ to $T=120$, corresponding to approximately 10 days of data in summer. We repeat the experiment with albedo as an additional input to evaluate whether the temporal context compensates for the lack of explicit surface reflectance information.

Additionally, a convolution residual network (ConvResNet), similar to that used in the SARAH-3 emulator \citep{schuurman_surface_2024}, is trained and used as a baseline for studying the benefits of the more complex attention-based architecture. More details about this baseline model are provided in Appendix \ref{appendix:baseline}.

\subsection{Emulator Characterization}
The best-performing context length from the the previous experiments is selected to define our HelioMont emulator, HeMu. To characterize HeMu's estimates accuracy, the deviation from ground truth is evaluated on the measurements of the SwissMetNet stations. Then HeMu's sensitivity to the input features is tested in two steps. First, individual features are removed from the training set and resulting SSR estimates are compared to the non-ablated HeMu outputs. Second, a permutation feature importance test \citep{fisher_all_2021}, also presented in \citep{schuurman_surface_2024}, is carried out to evaluate the sensitivity of the trained model to the inputs.

\subsection{Training Details}
All models are trained on the daylight (SZA > 80°) samples of the years 2015 and 2016, validated on 2017 data and tested on 2019 data. The adopted loss function is the mean-square-error (MSE), optimized using the AdamW method \citep{loshchilov_decoupled_2017} with an initial learning rate of $10^{-3}$, reduced by a factor 10 if no improvement of the loss function is measured for more than 1 epoch. The best checkpoint is selected based on the validation MSE, which is computed every 0.5 epoch. We train the transformers for 10 epochs. 

Following the hyper-parameters fine-tuning, the tokenization patch size is 3$\times$3$\times$1 and the attention window is 48$\times$48. Input images in their original size are thus split into 48$\times$48 subsets for inference. The 48$\times$48 SSR estimates are then reassembled to the original size. 
Models are trained on a single NVIDIA A100 GPU. 
The training time scales from 1.5 to 20 GPU-hours depending on $T$. The inference frame rate reaches approx. 10 frame/s, which is a speedup of approx. 10$\times$ with respect to the HelioMont inference time (Appendix \ref{appendix:benchmark}).

All presented results are for the test set. The error between the test output ($\hat{y}$) and the target (y) is measured using the mean bias error (MBE), the mean absolute error (MAE) and the root-mean-square error (RMSE): 
\begin{align*}
    \text{MBE} = \frac{1}{N} \sum_{i=1}^N(\hat{y}_i-y_i)
    \hspace{1cm}
    \text{MAE} = \frac{1}{N} \sum_{i=1}^N|\hat{y}_i-y_i|
    \hspace{1cm}
    \text{RMSE} = \sqrt{\frac{1}{N} \sum_{i=1}^N(\hat{y}_i-y_i)^2}\end{align*}

%% file: table1.tex
\begin{table}[h]
    \centering
    \caption{Overview of the emulator's training input and target variables}
    \begin{tabular}{|c|r|c|c|c|c|c|}
    \cline{1-7}
        & \multirow{2}{*}{\textbf{Variable}} &\multirow{2}{*}{\textbf{Source}} & \multicolumn{2}{|c|}{\textbf{Extent}}&\multicolumn{2}{|c|}{\textbf{Resolution}} \\
        \cline{4-7}
       &  & &\makecell{\textbf{Spatial}\\WGS84} & \makecell{\textbf{Temporal}\\UTC} & \makecell{\textbf{Spatial}\\WGS84}& \makecell{\textbf{Temporal}\\UTC}\\
     \cline{1-7}
        \rotatebox[origin=c]{270}{\textbf{TARGET}} & SSRGHI-No-Horizon& HelioMont \citep{castelli_heliomont_2014} & \makecell{Lon: [5.75°-10.75°]\\Lat: [45.75°-47.75°]}&[2015-2019]& \makecell{0.05°\\ $\approx$1.7km\\ (CH1903+)}&1h\\
        \cline{1-7}
         \multirow{15}{*}{\rotatebox[origin=c]{270}{\textbf{FEATURES}}} & High-Resolution Visible (HRV)&\multirow{9}{*}{MSG-SEVIRI \citep{schumann_msg_2002}} 
         &\multirow{9}{*}{\makecell{Lon: [5.75°-10.75°]\\Lat: [45.75°-47.75°] }}&\multirow{9}{*}{[2015-2019]}&0.05°&\multirow{9}{*}{1h}\\
         \cline{6-6}
         &Infrared 0.16$\mu$m (IR016)&&&&\multirow{9}{*}{1.5°}&\\
         &Infrared 0.39$\mu$m (IR039)&&&&&\\
         &Water Vapor 0.62$\mu$m (WV062)&&&&&\\
         &Water Vapor 0.73$\mu$m (WV073)&&&&&\\
         &Infrared 0.87$\mu$m (IR087)&&&&&\\
         &Infrared 0.97$\mu$m (IR097)&&&&&\\
         &Infrared 10.8$\mu$m (IR108)&&&&&\\
         &Infrared 12.0$\mu$m (IR120)&&&&&\\
         &Infrared 13.4$\mu$m (IR134)&&&&&\\
            \cline{2-7}
         &Elevation&\multirow{4}{*}{NASADEM \citep{opentopography_nasadem_2021}}&\multirow{3}{*}{\makecell{Lon: [5.75°-10.75°]\\Lat: [45.75°-47.75°] }}&\multirow{3}{*}{n.a.}&\multirow{4}{*}{0.05° }&\multirow{3}{*}{n.a.}\\
         &Slope&&&&&\\
         &Aspect&&&&&\\
         \cline{5-5}
         \cline{7-7}
         &$f_{corr}$&&&1 year&&1h\\
    \cline{2-7}
         &Solar Zenith Angle (SZA)&\multirow{2}{*}{(Grena, 2012) \citep{grena_five_2012}}&\multirow{2}{*}{\makecell{Lon: [5.75°-10.75°]\\Lat: [45.75°-47.75°] }}&\multirow{2}{*}{1 year}&\multirow{2}{*}{0.05° }&\multirow{2}{*}{1h}\\
         &Solar Azimuth Angle (SAA)&&&&&\\
     \cline{1-7}
    \end{tabular}
    \label{tab:featuresList}
\end{table}

%% file: 4_results.tex
\section{Results}
\label{sec:results}

\subsection{Temporal context experiment} 
\label{sec:results-context}
The temporal context size was increased from 1 to 120 steps, which extend the temporal context window into the past. For each temporal context size, two models were trained, including and not including ground albedo as input feature, respectively. 
The baseline model was also trained for both input feature scenarios.

\begin{figure}[H]
    \centering
    \includegraphics[width=\linewidth]{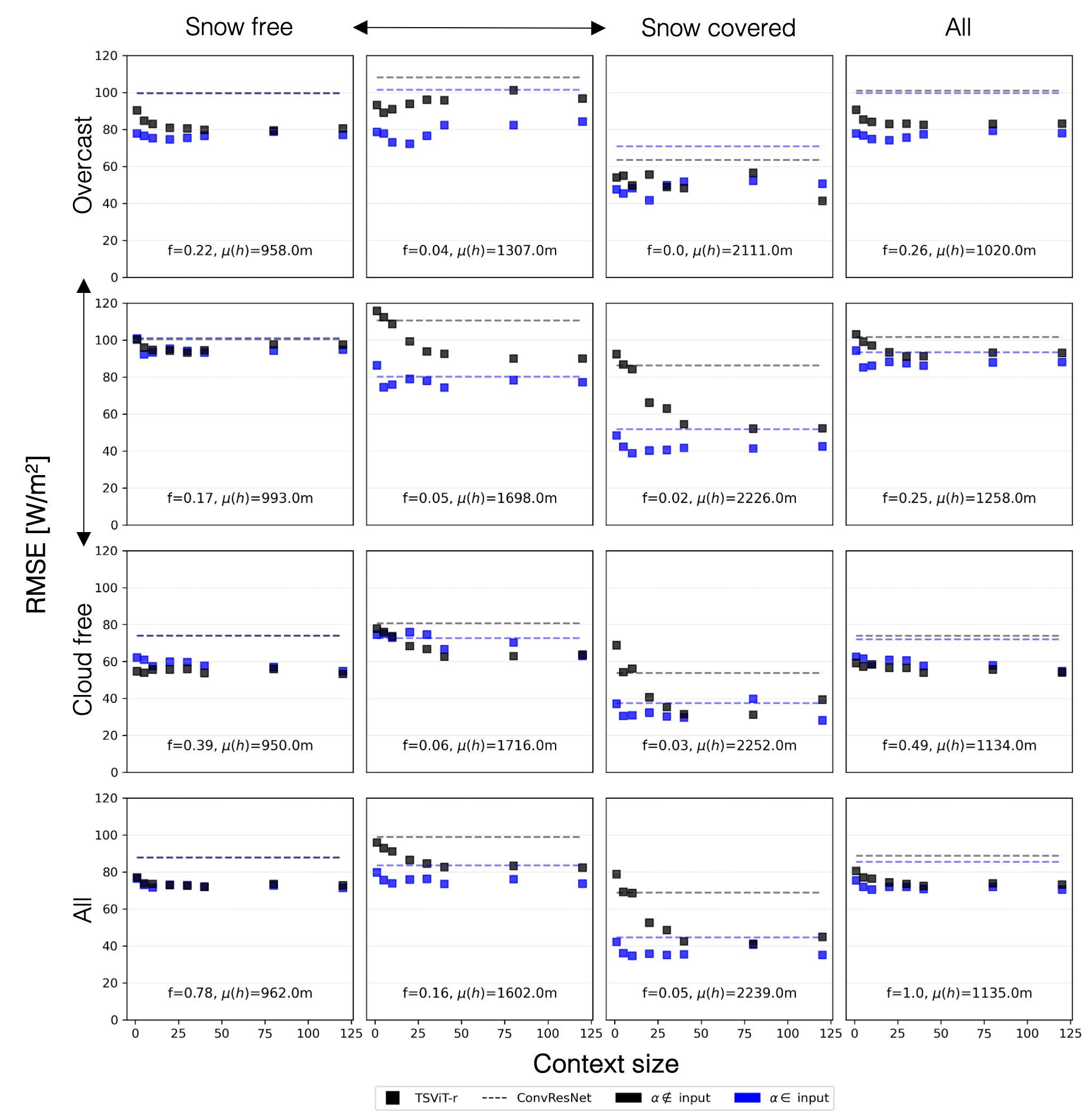}
    \caption{Root-mean-square errors (RMSE) of the emulator of the test set (2019) estimates, with and without albedo ($\alpha$) provided as input feature, are reported for temporal context lengths ranging from 1 to 120 time steps into the past. The RMSE of the baseline (context unaware ConvResNet model) is also provided for the two cases as dashed lines. The results are split by  $\alpha$ values (columns) and clear sky index $k_t^*$ (rows). The fraction ($f$) of the entire test data set that each subplot corresponds to, is given at the bottom of each subplot, along with this fraction's average ground elevation ($\mu(h)$).}
    \label{fig:1_rmse_seqlen}
\end{figure}

Figure~\ref{fig:1_rmse_seqlen} presents the RMSE values of the emulated SSR estimates across all test runs. To differentiate between the various atmospheric and ground reflectance conditions present in the test set, pixels within the study area are grouped by ground albedo ($\alpha$) and clear-sky index ($k_T^*$) values across the different subplots. In the first row only overcast pixels ($k_T^* \in [0,0.4[$) are considered. In the second row, mixed-sky clearness conditions are considered ($k_T^* \in [0.4,0.8[$) and in the third row, clear-sky conditions are considered ($k_T^* \in [0.8,1.2]$). The bottom-end row considers all values of $k_T^*$ indifferently. A similar split is applied column-wise for the values of  $\alpha$. In the left-most panel, the very dark, snow-free surfaces are shown ($\alpha \in [0,0.3[$), while the third panel presents results for the very bright, snow- or ice-covered surfaces ($\alpha \in [0.6,1[$). The right-most column contains all $\alpha$ values. Consequently the subplot in the lower right corner provides the results for the entire study area. 
On each subplot, the fraction ($f$) of the entire test set and the average ground elevation ($\mu(h)$) of this fraction's pixels are indicated. A large majority (78\%) of the test set contains very dark surfaces and the fraction of considered pixels decreases with increasing albedo value, to only 5\% for the upper-bound range. This is correlated to the mean ground elevation ($\mu(h)$), as snow- or ice-cover remains longer at higher elevations.\\

When albedo is not provided (black squares in Figure \ref{fig:1_rmse_seqlen}), increasing the temporal context length $T$ led to consistent overall improvements in RMSE, up to $T=40$. Beyond this context size, the estimates plateau at RMSE $\approx$ 70~W/m\textsuperscript{2}. 
The main observation of interest is that, in some cases, the gradual convergence of the RMSE values as the size of the context increases. 
In these cases, providing albedo as input feature improves the SSR estimates. Without context, the distance to the albedo non-informed models (TSViT-r and baseline) increases proportionally to $\alpha$ values, i.e. ground brightness. Increasing the context size allows to close this growing gap, suggesting that the information gained via temporal context can be substituted by explicit albedo inputs and vice-versa. A small sensitivity of the albedo-informed models to context size supports this suggestion. The curve drawn by the blue squares does decrease slightly for small context sizes, up to $T=10$, but remains constant then, suggesting that most of the gain observed for albedo non-informed models up to $T=40$ is substitutable by albedo information.

The cases for which these observations stand represent approx. a third (32\%) of the test data set. The strongest effects occur under mixed to clear-sky atmospheric conditions with very bright ground albedo, at an average elevation above 2200m asl. (middle-right column, second and third rows). To a lesser degree, context also impacts very dark pixels under overcast conditions, rather located in the study area's low-lands below 1000m asl. (top left corner). 
In most other cases, providing albedo as input does not lead to improvements in SSR estimates and so does the context size. As an exception, 4\% of the data set with medium $\alpha$ values and overcast conditions (top row, middle left column), do benefit from additional albedo information, but show a decrease in SSR estimates accuracy with increasing context size. 

In both cases (with and without albedo provided as input feature), the attention-based model (TSViT) outperformed the baseline convolution-based model (ConvResNet) across all sky clearness and ground albedo conditions. Notably, even without context ($T=1$), the TSViT architecture consistently yielded better SSR estimates than the baseline. However, in the cases (10\%) for which temporal context had the strongest impact, TSViT-r has similar or worse RMSE scores for $T=1$. 
The metrics for the entire study area (bottom-right subplot) are reported in Table~\ref{tab:rmse_seqlen} for all context sizes and both architectures.\\

\begin{figure}[h]
    \centering
    \includegraphics[width=1\linewidth]{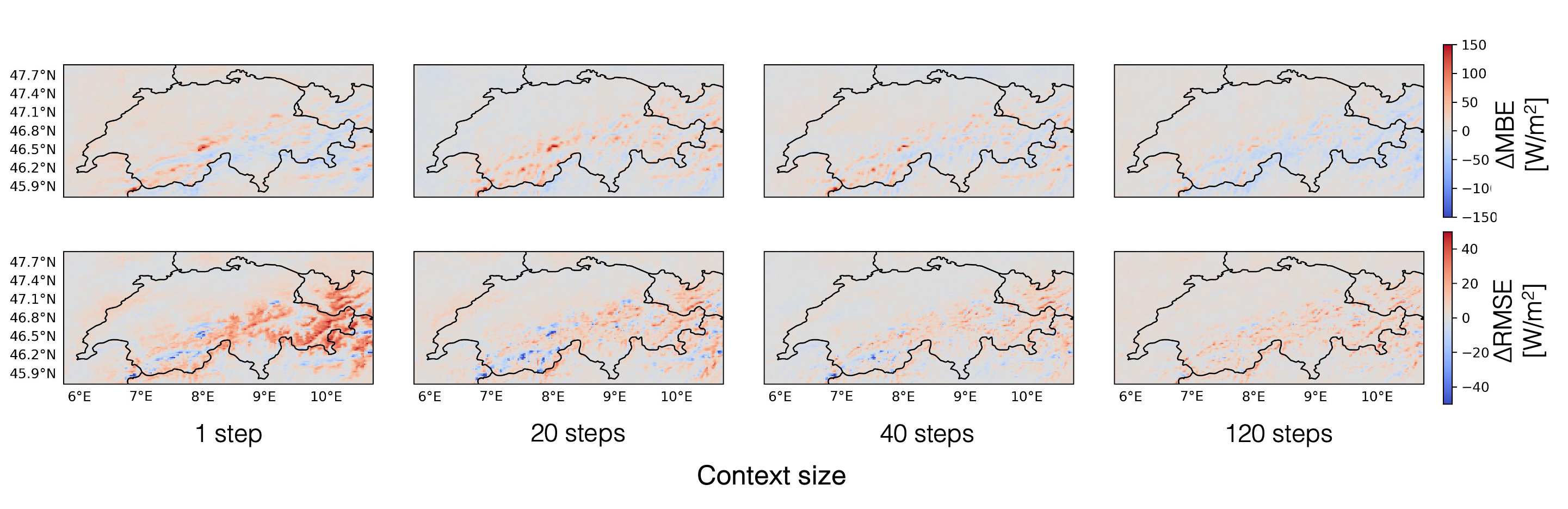}
    \caption{Difference in mean bias error (MBE, top row) and in root-mean-square error (RMSE, bottom row) of the emulator when trained with and without albedo ($\alpha$) as an input. The attenuation of the colors from left to right shows that the increase of the context size reduces the differences between the two models' outputs. The strongest remaining biases are located in the mountainous regions.}
    \label{fig:3_seqlen_map}
\end{figure}

Figure~\ref{fig:3_seqlen_map} presents maps of the differences in MBE ($\Delta$MBE) and in RMSE ($\Delta$RMSE) between albedo-informed and non-informed models (blue vs. black in Figure~\ref{fig:1_rmse_seqlen}), across increasing context lengths ($T = \{1, 20, 40, 120\}$).
As context increases, the spatial structure of the residual errors fades, indicating the convergence between the two models. The strongest reduction of RMSE occurs in the mountainous southern regions of the country, confirming that temporal context especially helps correcting the SSR overestimation of the model trained without background reflectance knowledge in snow-prone, high-reflectance, areas at higher elevations. However, the convergence is not uniform: some residual differences persist in topographically complex regions.

\input{table2}

\subsection{Characterization of the emulator}
As observed above, the gain in estimates accuracy stagnated beyond a context size of 40 time steps. Because a smaller context size also reduces the inference time, the HeMu emulator was run with a context size of 40.

\begin{figure}[h]
    \centering    
    \includegraphics[width=1\linewidth]{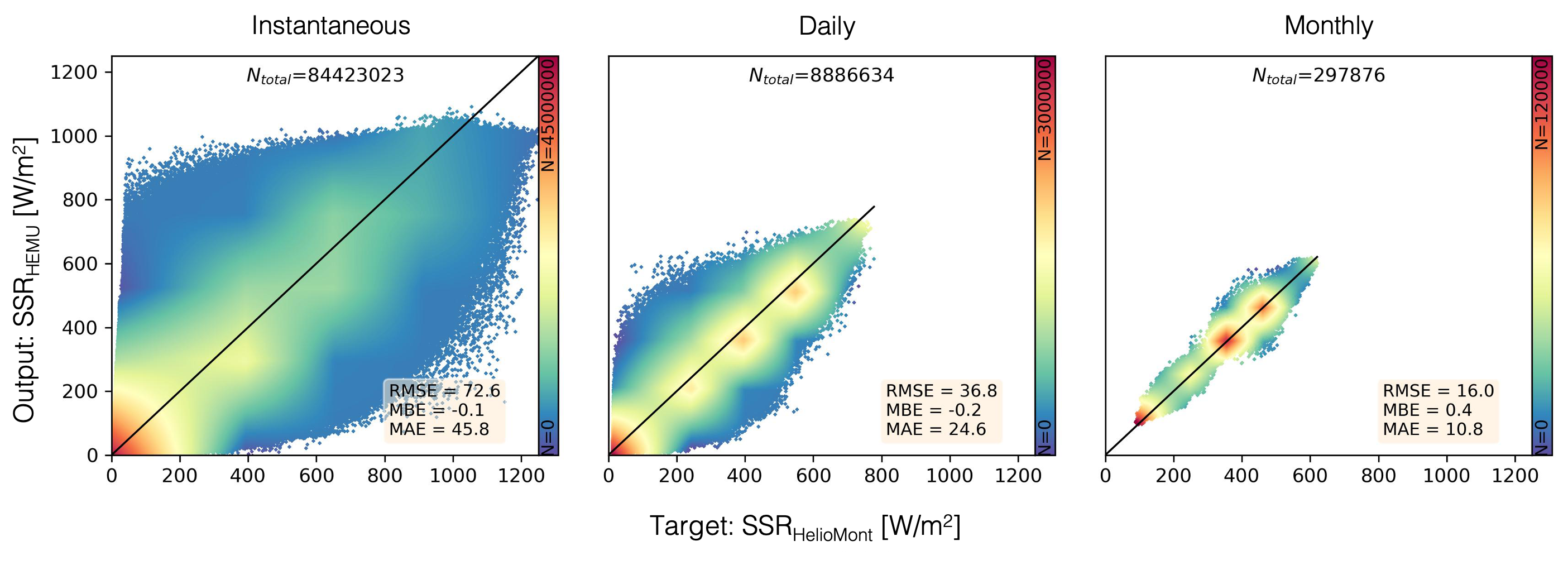} 
    \caption{Instantaneous (1/hour), daily and monthly HeMu's SSR estimates versus the target HelioMont SSR estimates over the test set (2019).}
    \label{fig:prod_scatter}
\end{figure}

Figure \ref{fig:prod_scatter} presents the output (HeMu SSR) versus target (HelioMont SSR) relationship for all points of the spatio-temporal extent studied, for instantaneous, daily and monthly aggregates. 

\begin{figure}[h]
    \centering    \includegraphics[width=1\linewidth]{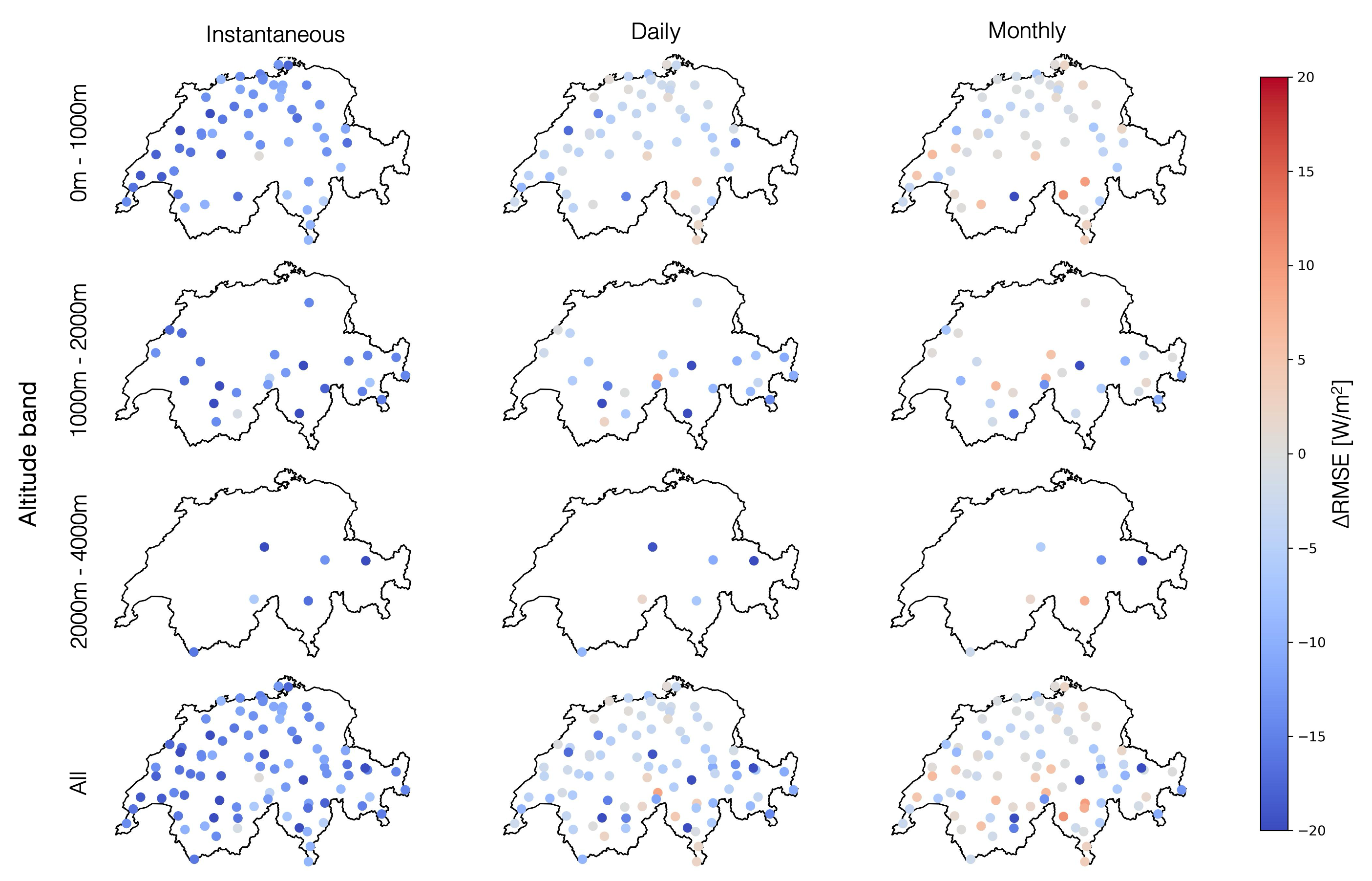} 
    \caption{Difference in RMSE error between SSR estimates of HeMu and HelioMont over the ground measurement of 87 stations of the SwissMetNet network \citep{noauthor_automatic_nodate}. Stations are split by the elevation band and the time resolution for which the metrics are computed. The corresponding metrics (RMSE, MBE and MAE) of both model are provided in table \ref{tab:smn_scores}.}
    \label{fig:prod_smn}
\end{figure}

The differences between HeMu's and HelioMont's RMSE for each SwissMetNet ground station is shown in Figure \ref{fig:prod_smn} split by elevation bands for instantaneous measurements, and daily / monthly aggregates. All scores are reported in Table \ref{tab:smn_scores}.
Because HeMu's instantenous estimates are rather negatively biased compared to HelioMont's estimates (Figure \ref{fig:prod_scatter}), especially for large values of GHI (>800 W/$m^2$), HeMu reduces the RMSE values at the ground stations (blue dots). HelioMont indeed shows an average positive bias over all stations (Table \ref{tab:smn_scores}). For daily and monthly aggregated values, the RMSE are more similar (pale blue and red tones). 
Over all stations of each elevation band, HeMu improves the RMSE values, while the mean bias error (MBE) is better or similar to the one obtained by HelioMont.

\input{table3}

\subsubsection{Training and inference sensitivity to the input features}
\begin{figure}[h]
    \centering
    \includegraphics[width=1\linewidth]{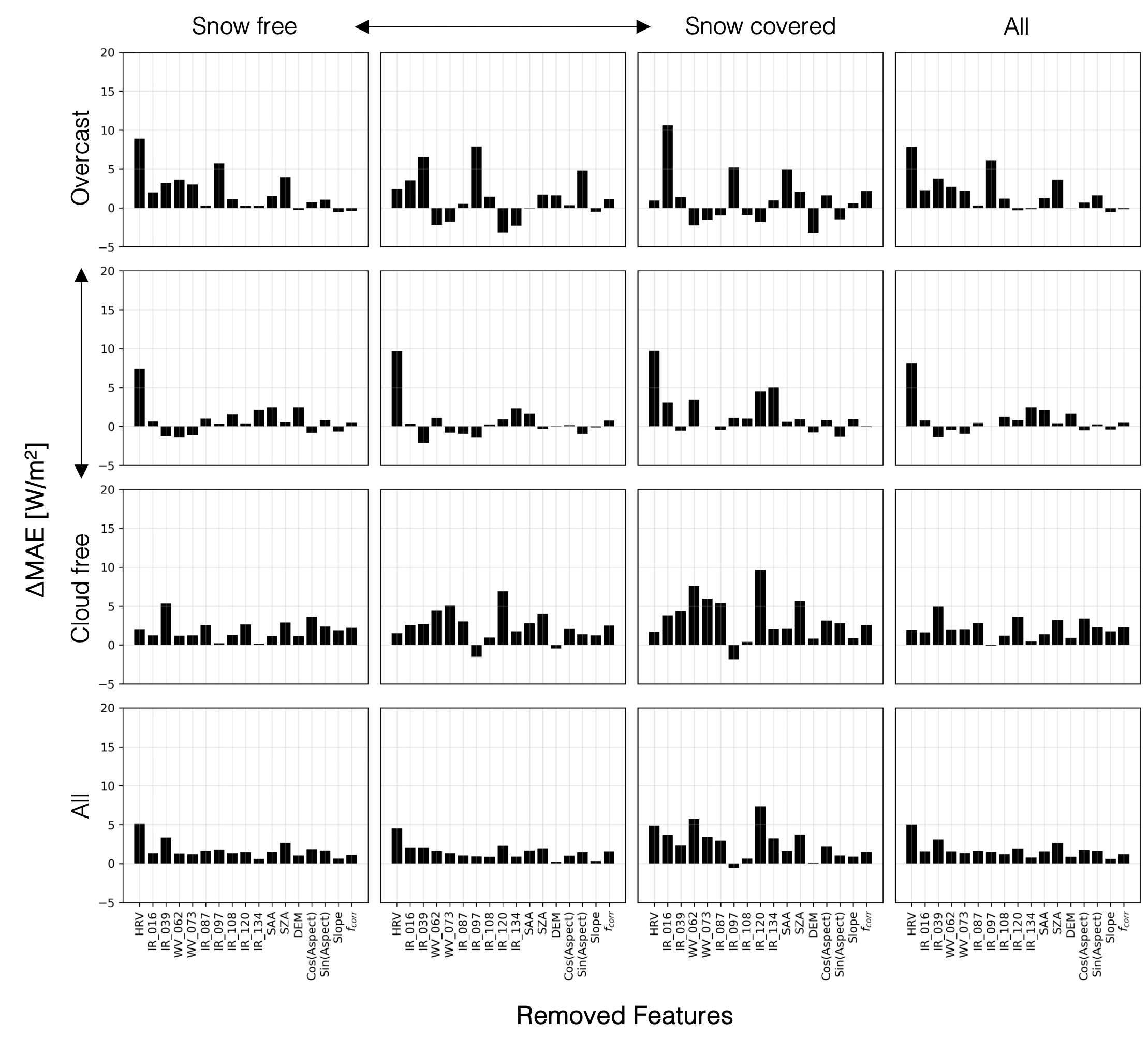} 
    \caption{Feature ablation study. Each input feature is alternately removed and the emulator fully retrained. The results are given in difference of the MAE scores (ablated - non-ablated).}
    \label{fig:ablation}
\end{figure}

\begin{figure}[h]
    \centering
    \includegraphics[width=1\linewidth]{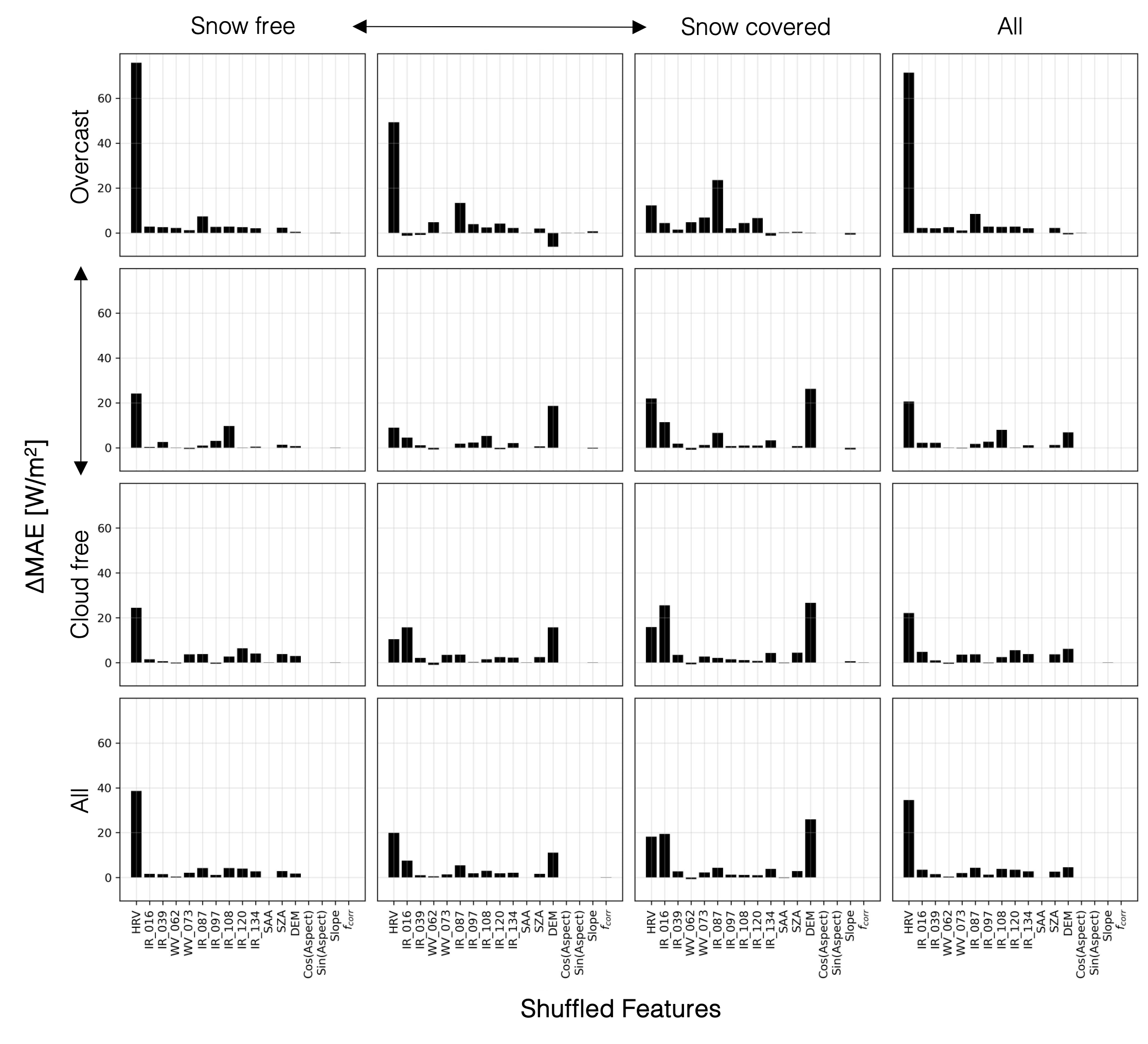} 
    \caption{Permutation feature importance. Each input feature is alternately randomized and corresponding inferences are compared to the reference, non-randomized, SSR estimates. The results are given in difference of the MAE scores (randomized - non-randomized).}
    \label{fig:permTest}
\end{figure}

To measure how the training of the model relies on the chosen input features, each feature was separately removed and the model re-trained. The inference of the resulting models was then compared to the reference, i.e. the model trained with all input features. Similarly, a permutation feature importance test was run to measure the sensitivity of a trained model to each input feature.
Figure \ref{fig:ablation} presents the difference in mean absolute error ($\Delta$MAE) of the SSR estimates for the sensitivity analysis at training time and Figure \ref{fig:permTest} at inference time.\\

Overall, a higher sensitivity was measured when the input features were shuffled at inference time. The most important feature, at both levels, is the high resolution visual (HRV) channel. At training time, the model actually compensates well the removal of single input features. The largest loss in MAE reaches $\approx$ 5 W/$m^2$ for HRV overall (bottom right subplot), up to $\approx$10 W/$m^2$ in overcast conditions (top row). 

At inference time, the permutation test shows an additional growing sensitivity to the infrared channels IR016 and IR087 as $\alpha$ increase. The former is more important under mixed to very clear sky conditions and the latter under overcast conditions. 
The importance of ground elevation information (DEM) is similarly growing with $\alpha$ values, under clear sky conditions. Other descriptors of the topography (aspect, slope), show very little to no impact in both experiments, witnessing redundancy or unnecessary information.

%% file: table2.tex
\begin{table}[ht]
    \centering
    \caption{Metrics from the context size experiment reported for the entire study area. The best scores are highlighted in bold. When albedo ($\alpha$) is provided as input, the RMSE and MAE stagnate around $\approx$70 W/$m^2$ and $\approx$45 W/$m^2$ respectively, with increasing context size. If $\alpha$ is removed from the input features, errors increase, eventually converging to those obtained when inputting albedo for longer context windows. The TSViT-r architecture outperforms the ConvResNet baseline.}
    \begin{tabular}{|cl|cccccccc|c|}
         \hline
         \multirow{2}{*}{$\alpha$ as input}&\multirow{2}{*}{Context Size} &\multicolumn{8}{c|}{TSViT-r}& Baseline\\
        && 1 & 5 & 10 & 20 & 30 & 40 & 80 & 120 & 1\\
         \hline
         \multirow{3}{*}{\xmark}&MBE [W/$m^2$]&-0.4&-1.5&-4.6&-5.0&0.8&-0.1&1.2&\textbf{-0.1}&-1.4\\
         &MAE [W/$m^2$]&52.2&49.9&49.6&47.8&47.2&\textbf{45.8}&46.8&46.7&59.6\\
         &RMSE [W/$m^2$]&80.7&77.1&76.6&74.5&73.7&\textbf{72.6}&73.9&73.3&88.8\\
         
         \hline
         \multirow{3}{*}{\cmark}&MBE [$W/m^2$]&-4.7&-7.5&-4.9&-7.7&-7.9&-2.2&-0.9&\textbf{-0.19}&11.3\\
         &MAE [W/$m^2$]&48.5&46.8&\textbf{44.6}&46.3&46.2&45.9&46.7&44.6&57.9\\
         &RMSE [W/$m^2$]&75.6&72.12&70.6&72.1&72.0&71.0&72.1&\textbf{70.6}&85.4\\
         \hline
         
    \end{tabular}

    \label{tab:rmse_seqlen}
\end{table}

%% file: table3.tex
\begin{table}[ht]
\centering
\caption{Metrics of HeMu's and HelioMont's instantaneous, daily and monthly estimates computed for the 87 ground stations measurements, split by elevation band.}
\begin{tabular}{|cc|lll|lll|lll|}
\hline
\multicolumn{1}{|c}{}        & \multicolumn{1}{c|}{} & \multicolumn{3}{c|}{Instantaneous [$W/m^2$]}                & \multicolumn{3}{c|}{Daily [$W/m^2$]}                     & \multicolumn{3}{c|}{Monthly [$W/m^2$]}                  \\
\multicolumn{1}{|c}{}        & \multicolumn{1}{c|}{} & MBE          & MAE            & RMSE           & MBE           & MAE           & RMSE          & MBE           & MAE           & RMSE          \\ 
\hline

\multirow{2}{*}{0m-1000m}    & HeMu                  & -0.6            & \textbf{94.8}    & \textbf{126.1} & -1.1          & \textbf{34.6} & \textbf{46.4} & -1.4            & \textbf{15.6} & \textbf{18.9}   \\
                             & HelioMont             & \textbf{-0.5} & 102.8          & 139.4          & \textbf{-0.1} & 36.7          & 49.8          & \textbf{-0.2} & 16.4          & 20.0          \\ 
\hline

\multirow{2}{*}{1000m-2000m} & HeMu                  & \textbf{0.5}   & \textbf{110.4}   & \textbf{146.6} & \textbf{3.3}  & \textbf{46.6} & \textbf{62.2} & \textbf{3.5}    & \textbf{22.5} & \textbf{28.1} \\
                             & HelioMont             & 5.3           & 120.6          & 161.8          & 8.8           & 50.7          & 70            & 9.1           & 25.9          & 32.2          \\
\hline

\multirow{2}{*}{2000m-4000m} & HeMu                  & -0.7            & \textbf{114.8} & \textbf{154.5} & -2.1          & \textbf{55.8} & \textbf{75.2} & -2.3          & \textbf{25.4} & 32.9          \\
                             & HelioMont             & \textbf{0.0}    & 127.2          & 173            & \textbf{-1.6} & 62.2          & 87.2          & \textbf{-1.6} & 41.1          & \textbf{32.1} \\ 
                             \hline
\multirow{2}{*}{0m-4000m}    & HeMu                  & -0.3          & \textbf{100.6}   & \textbf{134.0} & \textbf{0.1} & \textbf{39.5} & \textbf{52.9} & \textbf{0.0}   & \textbf{18.2}   & \textbf{22.5} \\
                             & HelioMont             & 1.2           & 109.6          & 148.2          & 2.4           & 42.5          & 58.2          & 2.3           & 20.2          & 25.0            \\ \hline
\end{tabular}
\label{tab:smn_scores}
\end{table}

%% file: 5_discussion.tex
\section{Discussion}
\label{sec:discussion}
\subsection{Temporal context leverages the inertia of ground albedo}
The experiments on context size showed that explicit albedo input could be replaced by an adequately long temporal context. It supported the hypothesis that temporal context enables implicit reconstruction of the background reflectance field without relying on snow nor cloud masks. 
However, the effects varied across different ground reflectance and atmospheric conditions nuancing the interpretation of the results. 
Specifically, the SSR estimates accuracy remained insensitive to the context size increase and to the explicit albedo input in some cases.
A clear impact was observed under two types of combinations: clear-sky and bright grounds or overcast and dark grounds. 

In both cases, there were large contrasts between dark ground and clouds or bright ground and clear sky. This indicates that the model extracts background reflectance through the higher stability of low (snow-free) and large (snow/ice) ground reflectance values in comparison to dynamic atmospheric conditions, to learn the background reflectance. It however remains unclear how much weight is given to a clear-sky observation over an overcast observation within the context window in that process.  
Future works will investigate the model's ability of compositing versus averaging, i.e. the capacity to extract clear-sky reflectance information in periods of strong dynamism of $\alpha$ and $k_T^*$ values (high frequency and large amplitude) as implemented in Heliomont, in contrast to a simpler statistic of the entire context window, similarly to the original Heliosat method. 

\subsection{The attention-based architecture requires fewer inductive priors}
The TSViT-r architecture outperformed the convolution-based baseline across all scenarios and proved to be notably robust to feature permutation and removal. While traditional convolutional networks rely on strong inductive biases like translation invariance and local spatial structure, TSViT-r's attention mechanism allows the model to flexibly capture both local and long-range dependencies in space and time. 

Our feature sensitivity analysis revealed that TSViT-r does not depend heavily on any single input at training time. The most impactful input was the high-resolution visible (HRV) channel, but its removal only increased the MAE score by ~5–10 W/m², and mostly under overcast conditions. A similar robustness was observed during permutation tests. Although the maximum increase in MAE reached approx. 80 $W/m^2$ for HRV in overcast conditions, the results contrast with earlier convolutional emulators, where sensitivity to features such as solar angles led to larger performance drops (MAE > 200 W/m²) when perturbed (e.g. \cite{schuurman_surface_2024}).

The reduced reliance on manually engineered input features makes the architecture more portable and less sensitive to errors in ancillary data like topographic variables or solar geometry. However, this also raises the question of optimal feature selection. Future work should investigate whether redundant or weakly contributing group of variables can be pruned to reduce model complexity without sacrificing accuracy.

\subsection{SSR estimates across simple and complex topography: limitations of an emulator}
When benchmarked against ground measurements, HeMu showed lower RMSE scores across all elevation bands, especially above 1000m. Notably, the improvement was most pronounced in instantaneous SSR estimates, where HeMu’s slight underestimation counterbalances HelioMont’s known overestimation bias at high irradiance levels. This suggests that HeMu may better handle clear-sky reflectance extremes without relying on fixed albedo composites.

Nevertheless, HeMu remains an emulator trained on HelioMont outputs, which means it inherits any systematic errors or domain limitations of the target algorithm. 
While HeMu improves on HelioMont’s estimates within Switzerland (the study area), especially in high-reflectance mountainous areas, its transferability to other regions remains to be validated.

%% file: 6_conclusion.tex
\section{Conclusion}
\label{sec:conclusion}
We proposed HeMu, a fully attention-based emulator for surface solar radiation retrieval that eliminates the need for explicit albedo input by leveraging temporal context. This approach allows the model to internally reconstruct background reflectance, especially over complex terrain with intermittent snow cover, where traditional compositing methods struggle.

Compared to convolutional baselines, HeMu achieves state-of-the-art accuracy and robustness, while requiring fewer hand-crafted input features. It matches HelioMont's performance on instantaneous, daily, and monthly SSR estimates, and even improves generalization in high-elevation areas. 

HeMu’s lightweight architecture and GPU-based inference offer a substantial speedup over the reference model.
This makes HeMu a promising candidate for operational solar energy applications, particularly in mountainous or snow-affected regions.

Future works will further investigate the role of topographical features and spatial resolution in the remaining biases towards a better model's generalization across ground elevation and complexity.

%% file: ack_data_cred.tex
\section*{Code availability}
The code for reproduction is available on: \url{https://github.com/frischwood/HeMu-dev.git}

\section*{Acknowledgment}
The authors were partly supported by InnoSuisse (Grant: 47985.1 IP-EE).

\section*{Declaration of Generative AI use}
Generative AI was used for language improvements. All suggested rephrasing were manually edited and checked. The authors take full responsibility of the actual content of this article. 

%% file: appendix.tex
\newpage
\section*{Appendix}
\section{Vision Transformers (ViTs)}
\label{app:vit_background}

A Vision Transformer (ViT) \citep{dosovitskiy_image_2021} is an adaptation of the Transformer architecture \citep{vaswani_attention_2023} to image data. It treats fixed-size image patches as tokens and applies the Transformer encoder to model long-range dependencies across the image.

Given an image $\mathbf{X} \in \mathbf{R}^{H \times W \times C}$, ViT splits it into $N = HW / P^2$ non-overlapping patches $\mathbf{x}_p \in \mathbf{R}^{P \times P \times C}$. Each patch is flattened and projected into a $d$-dimensional embedding using a linear layer, yielding a sequence of embeddings $\mathbf{e} \in \mathbf{R}^{N \times d}$. A learnable class token $\mathbf{x}_{\text{cls}} \in \mathbf{R}^{1 \times d}$ is prepended to form the input sequence $[\mathbf{x}_{\text{cls}}, \mathbf{e}] \in \mathbf{R}^{(N+1) \times d}$.

Since self-attention is permutation-invariant, ViTs require positional encodings to preserve spatial structure. A learnable position embedding $\mathbf{p} \in \mathbf{R}^{(N+1) \times d}$ is added to the input sequence, resulting in:
\[
\mathbf{z}_0 = [\mathbf{x}_{\text{cls}}, \mathbf{e}] + \mathbf{p}
\]

The Transformer encoder consists of $L$ blocks of alternating multi-head self-attention (MSA) and multi-layer perceptron (MLP) layers. Each block includes layer normalization (LN) and residual connections. Only the final state of the class token is used for classification.

While the original Transformer architecture includes both encoder and decoder modules (primarily for sequence-to-sequence tasks in NLP), ViTs use only the encoder, followed by an MLP head for classification.

For formal definitions of self-attention and of the Transformer architecture, we refer the reader to \citep{vaswani_attention_2023} and \citep{graves_generating_2014}.

\section{Temporo-Spatial Vision Transformer (TSViT)}
\label{app:TSVIt}

TSViT \citep{tarasiou_vits_2023} extends the ViT framework to satellite image time series (SITS), enabling spatiotemporal modeling for tasks such as crop classification and segmentation. The input is a SITS $\mathbf{x} \in \mathbf{R}^{T \times H \times W \times C}$ of $T$ time steps, where $H$, $W$ and $C$ are the height, width and channel dimensions of the SITS, repsectively.

\paragraph{Tokenization and Temporal Encoding.} TSViT first splits the input into 3D patches of shape $[t \times h \times w]$ and project it to the transformer's dimension $d$, yielding tokens of shape $\mathbf{R}^{N_T \times N_H \times N_W \times d}$ where:
\[
N_T = \left\lfloor \frac{T}{t} \right\rfloor,\quad N_H = \left\lfloor \frac{H}{h} \right\rfloor,\quad N_W = \left\lfloor \frac{W}{w} \right\rfloor
\]
In practice, TSViT uses $t=1$. The tokens are then reshaped into $N_H N_W$ temporal sequences of shape $[N_T \times d]$, one per spatial location. These sequences are passed to a temporal Transformer encoder.

The temporal encoder uses $K$ class tokens (one per target class) to improve class-specific representation. The $K$ output embeddings from the class tokens are reshaped into sequences of shape $[K \times N_H N_W \times d]$ for spatial encoding.

\paragraph{Spatial Encoding and Decoding.} The spatial encoder follows a standard ViT structure, operating on the sequences of class-token embeddings from the temporal encoder. TSViT provides two output heads:
\begin{itemize}
    \item A \textbf{segmentation head}, which ignores class tokens and projects the patch tokens back into the image domain via a linear layer, followed by reshaping into $[H \times W \times K]$.
    \item A \textbf{classification head}, which aggregates the final state of the class tokens to form a $[1 \times K]$ vector representing the class probabilities for the central image region.
\end{itemize}

This temporo-spatial factorization (temporal followed by spatial encoding) contrasts with many video models that use a spatio-temporal order. It is particularly well suited for SITS applications, where single-pixel time series carry significant semantic information (e.g., crop growth cycles) independent of neighboring pixels or object shapes.

\newpage
\section{HeMu's TSViT-based architecture}
\label{app:archi}
\begin{figure}[H]
    \centering
    \includegraphics[width=0.8\linewidth]{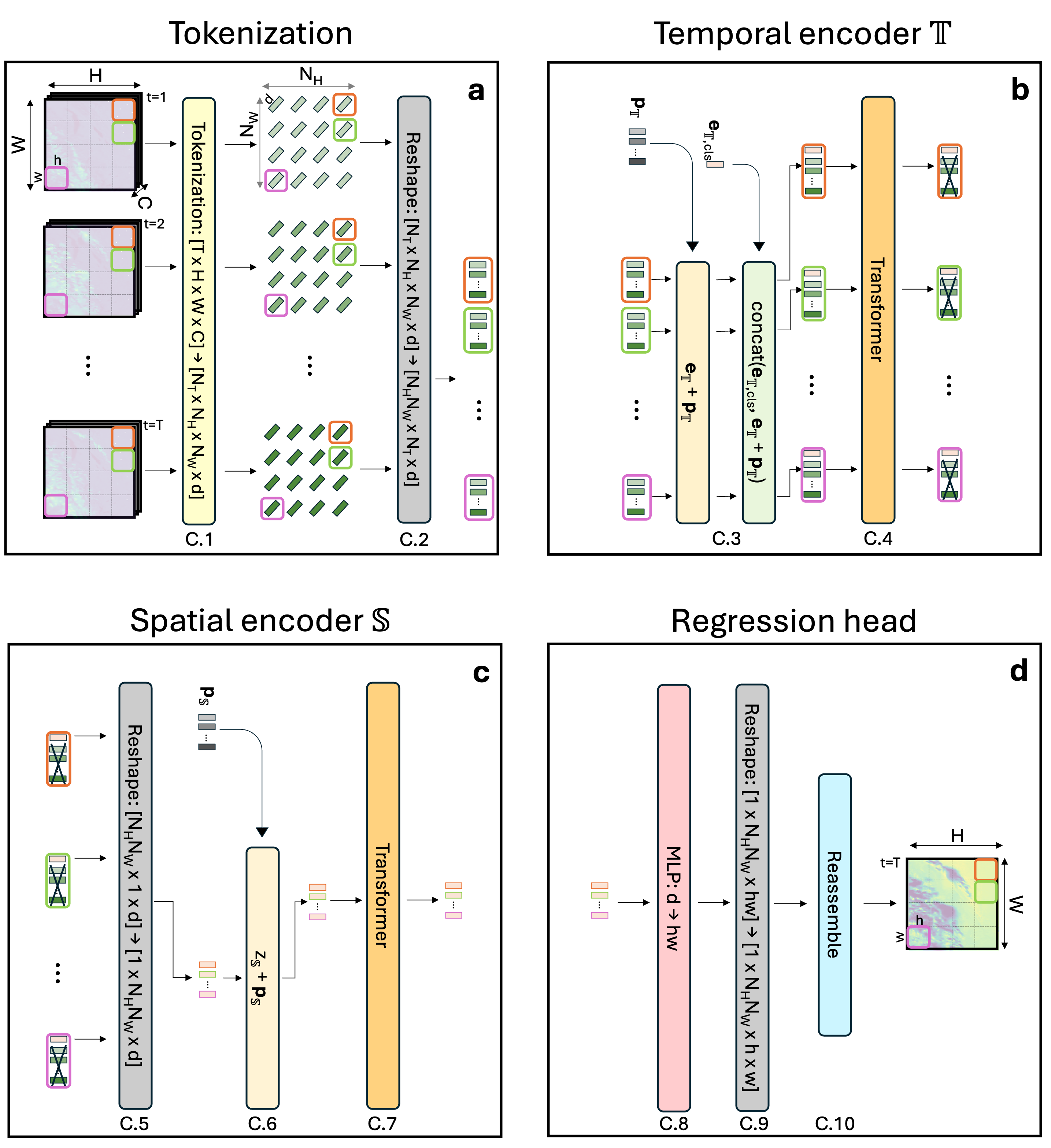}
    \caption{Architecture of the TSViT-r, adapted from \citep{tarasiou_vits_2023}.}
    \label{appendix:archi_model}
\end{figure}

\vspace{0.2cm}
\noindent\textbf{Tokenization and Temporal Encoder ($\mathbb{T}$)}
\begin{align}
\text{Tokenize: } & \mathbf{X} \in \mathbf{R}^{T \times H \times W \times C} \rightarrow \mathbf{e} \in \mathbf{R}^{N_T \times N_H \times N_W \times d} \\
\text{Reshape: } & \mathbf{e} \rightarrow \mathbf{e}_{\mathbb{T}} \in \mathbf{R}^{N_H N_W \times N_T \times d} \\
\text{Encode: } & \mathbf{z}_{\mathbb{T},0} \in \mathbf{R}^{N_H N_W \times (K + N_T)\times d} = [\mathbf{e}_{\mathbb{T},cls},\mathbf{e}_{\mathbb{T}} + \mathbf{p}_{\mathbb{T}}], \quad K=1\\
& \mathbf{z}_{\mathbb{T},l} = [\mathbf{z}^{cls}_{\mathbb{T},l},\mathbf{z}^{1}_{\mathbb{T},l},\mathbf{z}^{2}_{\mathbb{T},l} ...\mathbf{z}^{N_T}_{\mathbb{T},l}] =\text{MLP}_\mathbb{T}(\text{LN}(\text{MSA}_\mathbb{T}(\text{LN}(\mathbf{z}_{\mathbb{T},l-1})))),\quad l=1,\ldots,L_\mathbb{T} \quad (L_\mathbb{T}=8)
\end{align}

\vspace{0.2cm}
\noindent\textbf{Spatial Encoder ($\mathbb{S}$) and Decoder}
\begin{align}
\text{Reshape: } & \textbf{z}^{cls}_{\mathbb{T},L_t} \in \textbf{R}^{N_H N_W \times K \times d} \rightarrow \textbf{z}^{cls}_{\mathbb{S},L_\mathbb{T}} \in \textbf{R}^{K \times  N_H N_W \times d}\\
\text{Encode: } &\mathbf{z}_{\mathbb{S},0} \in \mathbf{R}^{K \times N_H N_W \times d} = \textbf{z}^{cls}_{\mathbb{S},L_t} + \mathbf{p}_\mathbb{S} \\
& \mathbf{z}_{\mathbb{S},l} = \text{MLP}_\mathbb{S}(\text{LN}(\text{MSA}_\mathbb{S}(\text{LN}(\mathbf{z}_{\mathbb{S},l-1})))),\quad l=1,\ldots,L_\mathbb{S} \quad(L_\mathbb{S}=4) \\
\text{Predict: } & \hat{\mathbf{y}}' = \text{MLP}_d(\mathbf{z}_{\mathbb{S},L_\mathbb{S}}) \in \mathbf{R}^{K \times N_H N_W \times hw} \\
\text{Reshape: } & \hat{\mathbf{y}}' \rightarrow \hat{\mathbf{y}} \in \mathbf{R}^{K \times N_H N_W \times h \times w}\\
\text{Reassemble: } & \hat{\mathbf{y}} \rightarrow \mathbf{R}^{H \times W \times K}
\end{align}

where, $\mathbf{X}$ is the time-series of length T of satellite images of size H$\times$W, which contains C channels including auxiliary data variables, and $\hat{\mathbf{y}}$ is the estimated SSR image of same size (H$\times$W) and K channels (here: K=1).

We use $L_\textit{T}$=8 blocks in the temporal encoder, $L_\textit{S}$=4 in the spatial encoder, and a 3$\times$3$\times$1 token patch size. The Transformers latent dimension is d=128. MLPs use two linear layers with GELU activations. The trainable parameters of the TSViT-r architecture sum up to 4345097 parameters.

\newpage
\section{Variability of the daylight data points over the year}
\label{appendix:dataset}
\begin{figure}[h]
    \centering
    \includegraphics[width=0.5\linewidth]{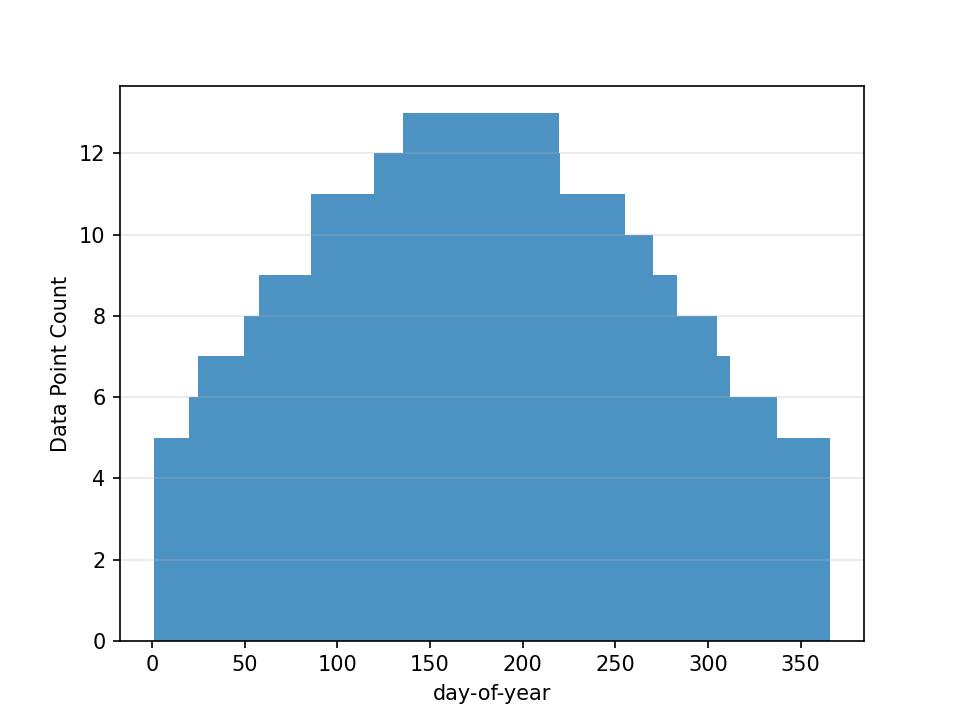}
    \caption{A filter on the solar zenith angle is applied to keep daylight data points only. The number of data points per day oscillates over the year between 5 in winter and 13 in summer.}
    \label{fig:hourCount}
\end{figure}

\section{Illumination and shadow correction factor}
\label{appendix:fcorr}
The illumination correction factor is referred to as the "correction factors for downwelling direct shortwave radiation" in HORAYZON \citep{steger_horayzon_2022}. When applied to solar radiation models that consider the ground as a flat surface, as HelioMont, it corrects for the self- and terrain-shading, the solar angle of incidence and the terrain deformation due to sloping surfaces. It is therefore an ideal feature to map the deformed satellite observations to the corrected undistorted SSR predictions of HelioMont. The factor is computed according to \citep{muller_grid-_2005}:
\begin{align*}
    f_{corr} = (\frac{1.0}{\textbf{h}\cdot\textbf{s}}) (\frac{1.0}{\textbf{h}\cdot\textbf{t}}) mask_{shadow} (\textbf{t}\cdot\textbf{s})
\end{align*}
where, for each pixel of the DEM, \textbf{h} is the normal of a horizontal surface, \textbf{s} is the sun position, \textbf{t} is the normal of the actual pixel oriented plane and $mask_{shadow}$ controls whether the considered pixel is shadowed or not (binary mask).

\newpage
\section{Baseline: ConvResNet}
\label{appendix:baseline}
\begin{figure}[h]
    \centering
    \includegraphics[width=1\linewidth]{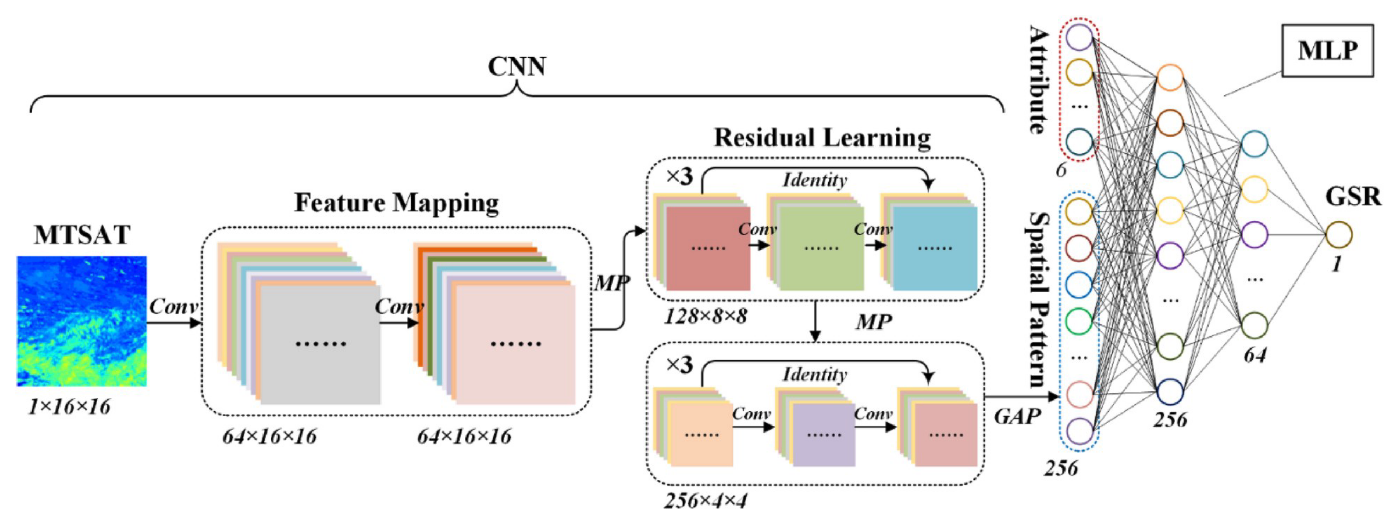}
    \caption{Convolutional residual network architecture proposed by \citep{jiang_deep_2019} and adopted for the SARAH-3 emulator \citep{schuurman_surface_2024} for SSR retrieval from satellite images.}
    \label{fig:baseline_archi}
\end{figure}

The convolutional residual network (ConvResNet), presented in \ref{fig:baseline_archi} and used as baseline in this article, was originally proposed in \citep{jiang_deep_2019}. This ConvResNet architecture does not take under consideration temporal context (context size = 1). Satellite input images are split in square patches which is ultimately the spatial context provided to infer the SSR estimate of the patch center pixel. \citep{schuurman_surface_2024} adopted this archtecture for their SARAH-3 emulator, only modifying the patch sizes from 16x16 to 15x15. Our baseline model adopts this last architecture but provides attributes as images as well, such that the upper part ("Attribute") of the MLP in Figure \ref{fig:baseline_archi} does not exist. 

\section{Context size effect split by ground elevation}
    \begin{figure}[h]
        \centering
        \includegraphics[width=1\linewidth]{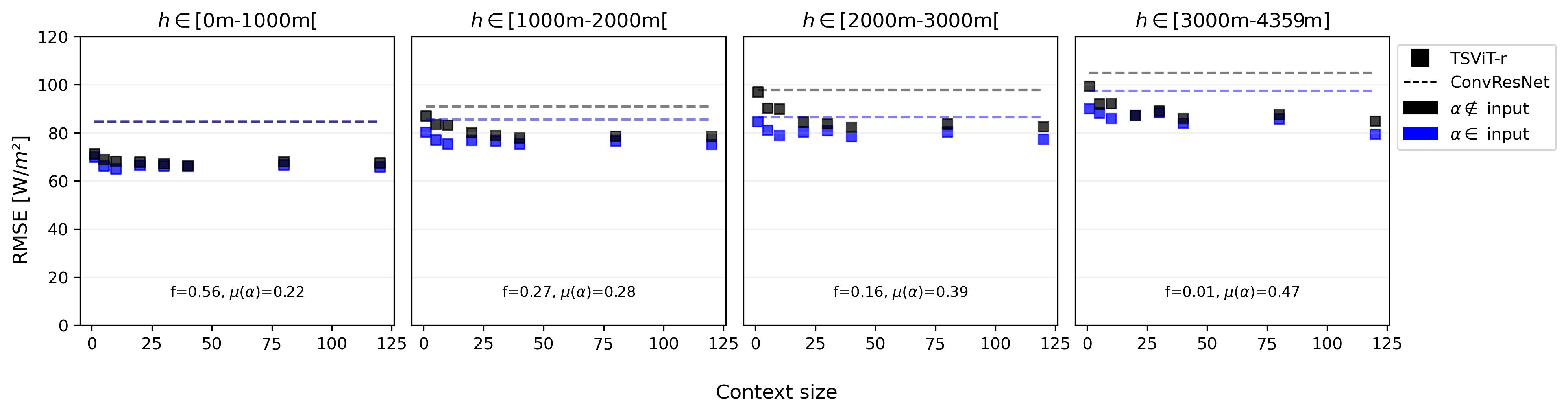}
        \caption{The results presented in Figure \ref{fig:1_rmse_seqlen} split by elevation band. For each band the fraction ($f$) and average ground albedo ($\mu(\alpha)$) are provided at the bottom of each subplot.}
        \label{fig:1_rmse_seqlen_alti}
    \end{figure}

\section{Benchmark for computational speedup}
\label{appendix:benchmark}
The original HelioMont set of algorithm is CPU-based. It is core-parallelized using OpenMP and can be node-parallelized using MPI. HeMu offers GPU-based inference using the Pytorch library and therefore scales up with the number of available GPUs. 
Both models are fundamentally different in their computational scheme which makes them hard to compare. We provide here the results of benchmark runs, aiming for a rough estimation of the computation time difference only. 
The benchmark is carried out over a single winter month for which HelioMont is known to be more computational intensive. HelioMont was run on a single 16-cores node and HeMu using a single NVIDIA A100 GPU (80GB). 
The average inference rate of HeMu was approx. 10 frame/s which is an increase of approx. 10$\times$ in comparison to HelioMont, excluding pre-processing.

\input{table4}

%% file: table4.tex
\begin{table}[h]
\centering
\caption{Results of the benchmark runs. Pre-processing and input and output (I/O) operations, depending on what is considered for each model take more computational time than the actual inference process. Here only the latter is provided to ease the comparison. Data loading, including transforms (e.g. normalization), is counted in the HeMu inference time, although it might be assimilated to I/O operations, because of its tight connection to the definition of ML-models.}
\begin{tabular}{|r|rc|rcc|}
\hline
                                & \multicolumn{2}{c|}{\textbf{HeMu} (GPU, NVIDIA A100)}     & \multicolumn{3}{c|}{\textbf{HelioMont} (CPU, 16-cores)}                                      \\
                                & Steps & Wall Time {[}s{]}    & Sub-steps                      & \multicolumn{2}{c|}{Wall Time {[}s{]}} \\
\hline

\multirow{13}{*}{Inference}    &   \multirow{13}{*}{\makecell{Forward pass \\(incl. data loading, \\transforms, etc.)}}   & \multirow{13}{*}{29} & Cloud Masking              & 13     & \multirow{13}{*}{310}  \\
                                &       &                      & Clear Sky Composite        & 218   &                           \\
                                &      &                      & Cloud Fraction             & 2     &                           \\
                                &       &                      & Cloud Top Height/Pressure  & 0      &                           \\
                                &       &                      & Land Surface Temperature   & 0         &                           \\
                                &       &                      & Surface Downward Rad.      & 0         &                           \\
                                &       &                      & Surface Outgoing Rad.      & 0         &                           \\
                                &       &                      & Free Tropospheric Humidity & 0         &                           \\
                                &       &                      & Albedo                     & 7
                                &                           \\
                                &       &                      & Surface Incoming Solar Rad & 5      &                           \\
                                &       &                      & openMP synchronizing          & 33     &                           \\
                                
                                &       &                      & Write Restart File         & 0      &                           \\
                                &       &                      & Exit                       & 0      & \\                         
\hline
\end{tabular}
\end{table}

%% file: main.bbl
\begin{thebibliography}{10}

\bibitem{iea_renewables_2024}
IEA.
\newblock Renewables 2024.
\newblock Technical report, IEA, Paris, 2024.
\newblock Licence: CC BY 4.0.

\bibitem{paletta_advances_2023}
Quentin Paletta, Guillermo Terrén-Serrano, Yuhao Nie, Binghui Li, Jacob Bieker, Wenqi Zhang, Laurent Dubus, Soumyabrata Dev, and Cong Feng.
\newblock Advances in solar forecasting: {Computer} vision with deep learning.
\newblock {\em Advances in Applied Energy}, 11:100150, September 2023.

\bibitem{sehrawat_solar_2023}
Neha Sehrawat, Sahil Vashisht, and Amritpal Singh.
\newblock Solar irradiance forecasting models using machine learning techniques and digital twin: {A} case study with comparison.
\newblock {\em International Journal of Intelligent Networks}, 4:90--102, 2023.

\bibitem{al-dahidi_enhancing_2024}
Sameer Al-Dahidi, Mohammad Alrbai, Hussein Alahmer, Bilal Rinchi, and Ali Alahmer.
\newblock Enhancing solar photovoltaic energy production prediction using diverse machine learning models tuned with the chimp optimization algorithm.
\newblock {\em Scientific Reports}, 14(1):18583, August 2024.

\bibitem{carpentieri_extending_2025}
A.~Carpentieri, D.~Folini, J.~Leinonen, and A.~Meyer.
\newblock Extending intraday solar forecast horizons with deep generative models.
\newblock {\em Applied Energy}, 377:124186, January 2025.

\bibitem{carpentieri_data-driven_2024}
Alberto Carpentieri, Jussi Leinonen, Jeff Adie, Boris Bonev, Doris Folini, and Farah Hariri.
\newblock Data-driven {Surface} {Solar} {Irradiance} {Estimation} using {Neural} {Operators} at {Global} {Scale}, November 2024.
\newblock arXiv:2411.08843 [physics].

\bibitem{perez_comparing_1997}
Richard Perez, Robert Seals, and Antoine Zelenka.
\newblock Comparing satellite remote sensing and ground network measurements for the production of site/time specific irradiance data.
\newblock {\em Solar Energy}, 60(2):89--96, February 1997.

\bibitem{ineichen_derivation_1999}
P.~Ineichen and R.~Perez.
\newblock Derivation of {Cloud} {Index} from {Geostationary} {Satellites} and {Application} to the {Production} of {Solar} {Irradiance} and {Daylight} {Illuminance} {Data}.
\newblock {\em Theoretical and Applied Climatology}, 64(1-2):119--130, October 1999.

\bibitem{huang_estimating_2019}
Guanghui Huang, Zhanqing Li, Xin Li, Shunlin Liang, Kun Yang, Dongdong Wang, and Yi~Zhang.
\newblock Estimating surface solar irradiance from satellites: {Past}, present, and future perspectives.
\newblock {\em Remote Sensing of Environment}, 233:111371, November 2019.

\bibitem{schroedter-homscheidt_surface_2022}
M.~Schroedter-Homscheidt, F.~Azam, J.~Betcke, N.~Hanrieder, M.~Lefèvre, L.~Saboret, and Y.‑M. Saint-Drenan.
\newblock Surface solar irradiation retrieval from {MSG}/{SEVIRI} based on {APOLLO} {Next} {Generation} and {HELIOSAT}‑4 methods.
\newblock {\em Meteorologische Zeitschrift}, 31(6):455--476, December 2022.

\bibitem{pfeifroth_sarah-3_2024}
Uwe Pfeifroth, Jaqueline Drücke, Steffen Kothe, Jörg Trentmann, Marc Schröder, and Rainer Hollmann.
\newblock {SARAH}-3 – satellite-based climate data records of surface solar radiation.
\newblock {\em Earth System Science Data}, 16(11):5243--5265, November 2024.

\bibitem{castelli_heliomont_2014}
M.~Castelli, R.~Stöckli, D.~Zardi, A.~Tetzlaff, J.E. Wagner, G.~Belluardo, M.~Zebisch, and M.~Petitta.
\newblock The {HelioMont} method for assessing solar irradiance over complex terrain: {Validation} and improvements.
\newblock {\em Remote Sensing of Environment}, 152:603--613, September 2014.

\bibitem{cano_method_1986}
D.~Cano, J.M. Monget, M.~Albuisson, H.~Guillard, N.~Regas, and L.~Wald.
\newblock A method for the determination of the global solar radiation from meteorological satellite data.
\newblock {\em Solar Energy}, 37(1):31--39, 1986.

\bibitem{emde_libradtran_2016}
Claudia Emde, Robert Buras-Schnell, Arve Kylling, Bernhard Mayer, Josef Gasteiger, Ulrich Hamann, Jonas Kylling, Bettina Richter, Christian Pause, Timothy Dowling, and Luca Bugliaro.
\newblock The {libRadtran} software package for radiative transfer calculations (version 2.0.1).
\newblock {\em Geoscientific Model Development}, 9(5):1647--1672, May 2016.

\bibitem{lefevre_mcclear_2013}
M.~Lefèvre, A.~Oumbe, P.~Blanc, B.~Espinar, B.~Gschwind, Z.~Qu, L.~Wald, M.~Schroedter-Homscheidt, C.~Hoyer-Klick, A.~Arola, A.~Benedetti, J.~W. Kaiser, and J.-J. Morcrette.
\newblock {McClear}: a new model estimating downwelling solar radiation at ground level in clear-sky conditions.
\newblock {\em Atmospheric Measurement Techniques}, 6(9):2403--2418, September 2013.

\bibitem{carpentieri_satellite-derived_2023}
A.~Carpentieri, D.~Folini, M.~Wild, L.~Vuilleumier, and A.~Meyer.
\newblock Satellite-derived solar radiation for intra-hour and intra-day applications: {Biases} and uncertainties by season and altitude.
\newblock {\em Solar Energy}, 255:274--284, May 2023.

\bibitem{jiang_deep_2019}
Hou Jiang, Ning Lu, Jun Qin, Wenjun Tang, and Ling Yao.
\newblock A deep learning algorithm to estimate hourly global solar radiation from geostationary satellite data.
\newblock {\em Renewable and Sustainable Energy Reviews}, 114:109327, October 2019.

\bibitem{lu_predicting_2023}
Yunbo Lu, Lunche Wang, Canming Zhu, Ling Zou, Ming Zhang, Lan Feng, and Qian Cao.
\newblock Predicting surface solar radiation using a hybrid radiative {Transfer}–{Machine} learning model.
\newblock {\em Renewable and Sustainable Energy Reviews}, 173:113105, March 2023.

\bibitem{gurel_state_2023}
Ali~Etem Gurel, Umit Agbulut, Huseyin Bakır, Alper Ergun, and Gokhan Yıldız.
\newblock A state of art review on estimation of solar radiation with various models.
\newblock {\em Heliyon}, 9(2):e13167, February 2023.

\bibitem{schuurman_surface_2024}
K.~R. Schuurman and A.~Meyer.
\newblock Surface solar radiation: {AI} satellite retrieval can outperform {Heliosat} and generalizes well to other climate zones, September 2024.
\newblock arXiv:2409.16316 [physics].

\bibitem{verbois_retrieval_2023}
Hadrien Verbois, Yves-Marie Saint-Drenan, Vadim Becquet, Benoit Gschwind, and Philippe Blanc.
\newblock Retrieval of surface solar irradiance from satellite imagery using machine learning: pitfalls and perspectives.
\newblock {\em Atmospheric Measurement Techniques}, 16(18):4165--4181, September 2023.

\bibitem{tarasiou_vits_2023}
Michail Tarasiou, Erik Chavez, and Stefanos Zafeiriou.
\newblock {ViTs} for {SITS}: {Vision} {Transformers} for {Satellite} {Image} {Time} {Series}, April 2023.
\newblock arXiv:2301.04944 [cs].

\bibitem{hammer_solar_2003}
Annette Hammer, Detlev Heinemann, Carsten Hoyer, Rolf Kuhlemann, Elke Lorenz, Richard Müller, and Hans~Georg Beyer.
\newblock Solar energy assessment using remote sensing technologies.
\newblock {\em Remote Sensing of Environment}, 86(3):423--432, August 2003.

\bibitem{bird_simple_1986}
Richard~E. Bird and Carol Riordan.
\newblock Simple {Solar} {Spectral} {Model} for {Direct} and {Diffuse} {Irradiance} on {Horizontal} and {Tilted} {Planes} at the {Earth}'s {Surface} for {Cloudless} {Atmospheres}.
\newblock {\em Journal of Climate and Applied Meteorology}, 25(1):87--97, January 1986.

\bibitem{muneer_solar_1990}
T.~Muneer.
\newblock Solar radiation model for {Europe}.
\newblock {\em Building Services Engineering Research and Technology}, 11(4):153--163, November 1990.

\bibitem{hay_calculating_1993}
John~E. Hay.
\newblock Calculating solar radiation for inclined surfaces: {Practical} approaches.
\newblock {\em Renewable Energy}, 3(4):373--380, June 1993.

\bibitem{tan_illumination_2010}
Bin Tan, Robert Wolfe, Jeffrey Masek, Feng Gao, and Eric~F. Vermote.
\newblock An illumination correction algorithm on {Landsat}-{TM} data.
\newblock In {\em 2010 {IEEE} {International} {Geoscience} and {Remote} {Sensing} {Symposium}}, pages 1964--1967, Honolulu, HI, USA, July 2010. IEEE.

\bibitem{von_rutte_how_2021}
F.~von Rütte, A.~Kahl, J.~Rohrer, and M.~Lehning.
\newblock How {Forward}-{Scattering} {Snow} and {Terrain} {Change} the {Alpine} {Radiation} {Balance} {With} {Application} to {Solar} {Panels}.
\newblock {\em Journal of Geophysical Research: Atmospheres}, 126(15), 2021.

\bibitem{khlopenkov_sparc_2007}
Konstantin~V. Khlopenkov and Alexander~P. Trishchenko.
\newblock {SPARC}: {New} {Cloud}, {Snow}, and {Cloud} {Shadow} {Detection} {Scheme} for {Historical} 1-km {AVHHR} {Data} over {Canada}.
\newblock {\em Journal of Atmospheric and Oceanic Technology}, 24(3):322--343, March 2007.

\bibitem{opentopography_shuttle_2013}
{OpenTopography}.
\newblock Shuttle {Radar} {Topography} {Mission} ({SRTM}) {Global}, 2013.

\bibitem{stockli_heliomont_2022}
Reto Stöckli.
\newblock The {HelioMont} {Surface} {Solar} {Radiation} {Processing} (2022 {Version}).
\newblock Scientific~93, MeteoSwiss, December 2022.

\bibitem{schumann_msg_2002}
W~Schumann, H~Stark, K~McMullan, D~Aminou, and H-J Luhmann.
\newblock {MSG} {Project}, {ESA} {Directorate} of {Earth} {Observation}, {ESTEC}, {Noordwijk}, {The} {Netherlands}.
\newblock 2002.

\bibitem{grena_five_2012}
Roberto Grena.
\newblock Five new algorithms for the computation of sun position from 2010 to 2110.
\newblock {\em Solar Energy}, 86(5):1323--1337, May 2012.

\bibitem{muller_grid-_2005}
Mathias~D. Müller and Dieter Scherer.
\newblock A {Grid}- and {Subgrid}-{Scale} {Radiation} {Parameterization} of {Topographic} {Effects} for {Mesoscale} {Weather} {Forecast} {Models}.
\newblock {\em Monthly Weather Review}, 133(6):1431--1442, June 2005.

\bibitem{opentopography_nasadem_2021}
{OpenTopography}.
\newblock {NASADEM} {Global} 1 arc-second {Digital} {Elevation} {Model}, 2021.

\bibitem{steger_horayzon_2022}
Christian~R. Steger, Benjamin Steger, and Christoph Schär.
\newblock {HORAYZON} v1.2: an efficient and flexible ray-tracing algorithm to compute horizon and sky view factor.
\newblock {\em Geoscientific Model Development}, 15(17):6817--6840, September 2022.

\bibitem{noauthor_automatic_nodate}
Automatic measurement network - {MeteoSwiss}.

\bibitem{martin_raspaud_pytrollsatpy_2025}
Martin Raspaud, David Hoese, Panu Lahtinen, Gerrit Holl, Simon Proud, Stephan Finkensieper, Andrea Meraner, Adam Dybbroe, Johan Strandgren, yukaribbba, Joleen Feltz, Sauli Joro, BENR0, Xin Zhang, Pierre de~Buyl, Gionata Ghiggi, William Roberts, Youva, Lars Ørum Rasmussen, Olivier Samain, mherbertson, Jorge Humberto~Bravo Méndez, Yufei Zhu, Isotr0py, seenno, rdaruwala, ClementLaplace, and bkremmli.
\newblock pytroll/satpy: {Version} 0.54.0 (2025/01/20), January 2025.

\bibitem{fisher_all_2021}
Aaron Fisher, Cynthia Rudin, and Francesca Dominici.
\newblock All {Models} are {Wrong}, but {Many} are {Useful}: {Learning} a {Variable}’s {Importance} by {Studying} an {Entire} {Class} of {Prediction} {Models} {Simultaneously}.
\newblock 2021.

\bibitem{loshchilov_decoupled_2017}
Ilya Loshchilov and Frank Hutter.
\newblock Decoupled {Weight} {Decay} {Regularization}, 2017.
\newblock Version Number: 3.

\bibitem{dosovitskiy_image_2021}
Alexey Dosovitskiy, Lucas Beyer, Alexander Kolesnikov, Dirk Weissenborn, Xiaohua Zhai, Thomas Unterthiner, Mostafa Dehghani, Matthias Minderer, Georg Heigold, Sylvain Gelly, Jakob Uszkoreit, and Neil Houlsby.
\newblock An {Image} is {Worth} 16x16 {Words}: {Transformers} for {Image} {Recognition} at {Scale}, June 2021.
\newblock arXiv:2010.11929 [cs].

\bibitem{vaswani_attention_2023}
Ashish Vaswani, Noam Shazeer, Niki Parmar, Jakob Uszkoreit, Llion Jones, Aidan~N. Gomez, Lukasz Kaiser, and Illia Polosukhin.
\newblock Attention {Is} {All} {You} {Need}, August 2023.
\newblock arXiv:1706.03762 [cs].

\bibitem{graves_generating_2014}
Alex Graves.
\newblock Generating {Sequences} {With} {Recurrent} {Neural} {Networks}, June 2014.
\newblock arXiv:1308.0850 [cs].

\end{thebibliography}
